\PassOptionsToPackage{svgnames,table,xcdraw}{xcolor}
\PassOptionsToPackage{most}{tcolorbox}
\PassOptionsToPackage{colorlinks=true,citecolor=lightblue,linkcolor=lightblue,urlcolor=lightblue}{hyperref}

\documentclass[11pt,letterpaper]{mystyle}

\usepackage[comma,authoryear,compress]{natbib}
\usepackage{xcolor}
\definecolor{lightblue}{rgb}{0.22,0.45,0.70}
\definecolor{highlight}{gray}{0.92}

\makeatletter
\@ifpackageloaded{hyperref}{}{%
  \usepackage{hyperref}%
}
\makeatother

\usepackage[all]{hypcap}
\usepackage{cleveref}
\DeclareUnicodeCharacter{FFFD}{\textemdash}

\usepackage{microtype}
\usepackage{graphicx}
\usepackage{subcaption}
\usepackage{booktabs}
\usepackage{amsmath,amssymb,amsfonts}
\usepackage{algorithmic}
\usepackage{textcomp}
\usepackage{makecell}
\usepackage{multirow}
\usepackage{diagbox}
\usepackage{balance}
\usepackage[flushleft]{threeparttable}
\usepackage[utf8]{inputenc} 
\usepackage[T1]{fontenc}    
\usepackage{url}            
\usepackage{nicefrac}      
\usepackage{listings}
\usepackage{adjustbox}
\usepackage{algorithm}
\usepackage{enumitem}
\usepackage{pifont}
\usepackage{colortbl}
\usepackage{authblk}
\usepackage{wrapfig}

\makeatletter
\@ifundefined{theorem}{}{}
\@ifundefined{corollary}{}{}
\makeatother

\newcommand{\toolname}[0]{\textbf{\textsc{UI-Oceanus}}}

\title{\toolname{}: Scaling GUI Agents with Synthetic Environmental Dynamics}

\author[1,2]{Mengzhou Wu}
\author[1,2]{Yuzhe Guo}
\author[1,2]{Yuan Cao}
\author[3]{Haochuan Lu}
\author[3]{Songhe Zhu}
\author[3]{Pingzhe Qu}
\author[3]{Xin Chen}
\author[3]{Kang Qin}
\author[3]{Zhongpu Wang}
\author[3]{Xiaode Zhang}
\author[3]{Xinyi Wang}
\author[3]{Wei Dai}
\author[3]{Gang Cao}
\author[3]{Yuetang Deng}
\author[3]{\authorcr Zhi Gong}
\author[1,2]{Dezhi Ran\textsuperscript{$\dagger$,$*$,}}
\author[5]{Linyi Li}
\author[4]{Wei Yang}
\author[1,2,6,7]{Tao Xie\textsuperscript{$*$,}}

\affil[1]{Key Lab of HCST (PKU), MOE; School of Computer Science, Peking University, China}
\makeatletter
\renewcommand\AB@affilsepx{\protect\\\protect\Affilfont}
\makeatother
\affil[2]{Beijing Tongming Lake Information Technology Application Innovation Center, China}
\makeatletter
\renewcommand\AB@affilsepx{\quad\protect\Affilfont} 
\makeatother
\affil[3]{WeChat, Tencent Inc., China}
\makeatletter
\renewcommand\AB@affilsepx{\quad\protect\Affilfont}
\makeatother
\affil[4]{University of Texas at Dallas, USA}
\makeatletter
\renewcommand\AB@affilsepx{\protect\\\protect\Affilfont}
\makeatother
\affil[5]{Simon Fraser University, Canada}
\makeatletter
\renewcommand\AB@affilsepx{\protect\\\protect\Affilfont}
\makeatother
\affil[6]{Fudan University Institute of Systems for Advanced Computing, China}
\makeatletter
\renewcommand\AB@affilsepx{\protect\\\protect\Affilfont}
\makeatother
\affil[7]{Shanghai Institute of Systems for Open Computing, China}
\makeatletter
\renewcommand\AB@affilsepx{\protect\\\protect\Affilfont} 
\makeatother
\affil[ ]{\rule{0pt}{2.5em}\small $^\dagger$Project Leader \quad $^*$Corresponding Authors}

\begin{document}

\begin{abstract}
\noindent Scaling generalist GUI agents is hindered by the data scalability bottleneck of expensive human demonstrations and the ``distillation ceiling'' of synthetic teacher supervision.
To transcend these limitations, we propose \toolname{}, a framework that shifts the learning focus from mimicking high-level trajectories to mastering interaction physics via ground-truth environmental feedback.
Through a systematic investigation of self-supervised objectives, we identify that forward dynamics, defined as the generative prediction of future interface states, acts as the primary driver for scalability and significantly outweighs inverse inference.
\toolname{} leverages this insight by converting low-cost autonomous exploration, which is verified directly by system execution, into high-density generative supervision to construct a robust internal world model.
Experimental evaluations across a series of models demonstrate the decisive superiority of our approach: models utilizing Continual Pre-Training (CPT) on synthetic dynamics outperform non-CPT baselines with an average success rate improvement of 7\% on offline benchmarks, which amplifies to a 16.8\% gain in real-world online navigation.
Furthermore, we observe that navigation performance scales with synthetic data volume. These results confirm that grounding agents in forward predictive modeling offers a superior pathway to scalable GUI automation with robust cross-domain adaptability and compositional generalization.
\end{abstract}

\maketitle

\section{Introduction}
\label{sec:intro}

Generalist GUI agents~\cite{qin2025ui, wang2025ui, wu2024atlas, xu2024aguvis, hong2024cogagent} traditionally rely on Behavioral Cloning (BC)~\cite{torabi2018behavioral, rawles2023androidinthewild, zhang2024android, lu2024gui, chen2024guicourse} to map visual observations to actions.
However, acquiring trajectory-level data is prohibitively expensive. Unlike static image-text pairs, valid execution traces require long-horizon sequences where a single error can invalidate the entire trajectory. This sensitivity makes synthesizing reliable data difficult and necessitates labor-intensive human annotation~\cite{zhang2025progrm, rawles2023androidinthewild, lu2024gui, li2024effects}, creating a major bottleneck for data scalability.

To bypass this bottleneck, recent approaches~\cite{tang2025magicgui, wu2024atlas} have adopted distillation-based Continual Pre-training (CPT), distilling knowledge from foundation models via static objectives like UI captioning~\cite{li2020widget} or grounding~\cite{cheng2024seeclick}.
Critically, these methods suffer from a ``distillation ceiling''~\cite{gou2021knowledge}: they enhance the agent's \emph{static semantic understanding} but fail to capture the \emph{temporal dynamics} of the environment.
For instance, an agent might correctly identify a ``submit'' button (semantics) but stubbornly assume that clicking it leads to submission, failing to anticipate that an incomplete form will actually trigger a validation error (dynamics).
Lacking such grounded interaction physics, these agents remain bound by static semantic priors, unable to adapt when dynamic environmental states contradict their shallow heuristics.
Therefore, there is a pressing need for a scalable CPT objective that derives supervision directly from autonomous environmental feedback.

\begin{wrapfigure}{r}{0.5\columnwidth} 
    \centering
    \includegraphics[width=\linewidth]{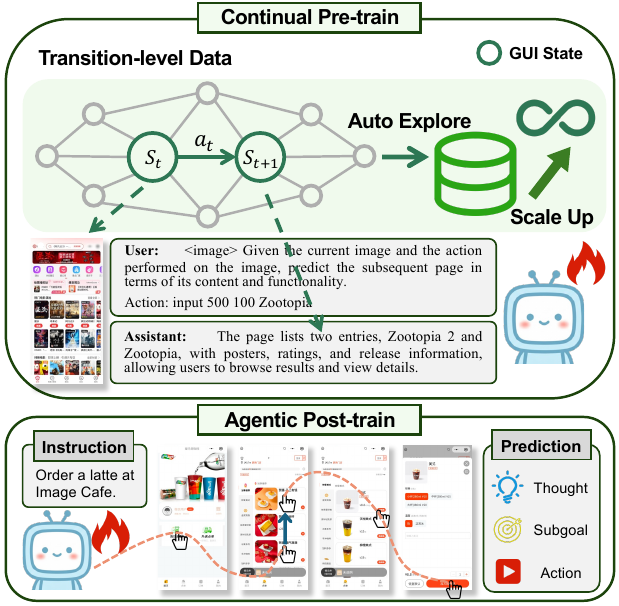} 
    \caption{\textbf{Constructing Generalist GUI Agents via Scalable World Model Learning.} (Top) We first establish a robust physical foundation by learning a forward dynamics world model from massive, autonomously explored transitions. (Bottom) We then leverage this internalized world model to instantiate a generalist GUI agent through agentic post-training.}
    \label{fig:teaser}
\end{wrapfigure}

To address these limitations, we propose \toolname{}, a self-supervised training framework designed to equip agents with a robust GUI world model (Figure~\ref{fig:teaser}).
The core insight in \toolname{} is that the underlying set of atomic transitions (i.e., single-step observation-action-outcome tuples) in the GUI environment is both relatively \emph{finite} and inherently \emph{self-supervised}, providing a natural task-agnostic scaling dimension for CPT.
The \emph{finite} nature ensures that the environmental dynamics can be efficiently explored without an expert policy model or explicit task goal to reach high coverage.
The \emph{self-supervised} nature enables the environment to act as a rigorous verifier where the observed transition following the action serves as a ground-truth label of consequence.
Both features enable agents to utilize task-agnostic exploration heuristics~\cite{wu2024skill, ran2023badge, gu2019practical, google2021android, su2017guided} to efficiently harvest vast quantities of these atomic transitions.
Crucially, these atomic transitions ground the learning process in mastering the \emph{fundamental mechanics} of the interface first.
This effectively separates the acquisition of interaction physics from the execution of user intent: the agent learns ``what is possible'' (dynamics) from massive synthetic data before learning ``what is desirable'' (intent) from scarce expert demonstrations.

To realize this framework, we develop a scalable data engine capable of converting low-cost autonomous exploration into high-density supervision.
We systematically investigate the spectrum of environmental dynamics and uncover a critical insight: forward dynamics, defined as the generative prediction of future interface states, serves as the key objective enabling performance scaling with data volume, significantly outperforming inverse dynamics and backward dynamics.
Consequently, \toolname{} prioritizes \emph{forward dynamics prediction} as the generative objective~\cite{ha2018world, hafner2019dream}.
Technically, we capture raw interaction traces, isolate unique and meaningful atomic transitions via strict hashing filtering and semantic filtering, and employ a Visual-Language Model~(VLM) to synthesize precise descriptions of environmental changes.
Finally, we implement a two-stage training strategy: the model is first continually pre-trained on a mixture of GUI dynamics, grounding, and general data to establish a robust world model, and subsequently specialized via agentic post-training to align this physical intuition with complex instruction following.

\begin{figure*}[!t] 
    \centering
    \includegraphics[width=0.95\textwidth]{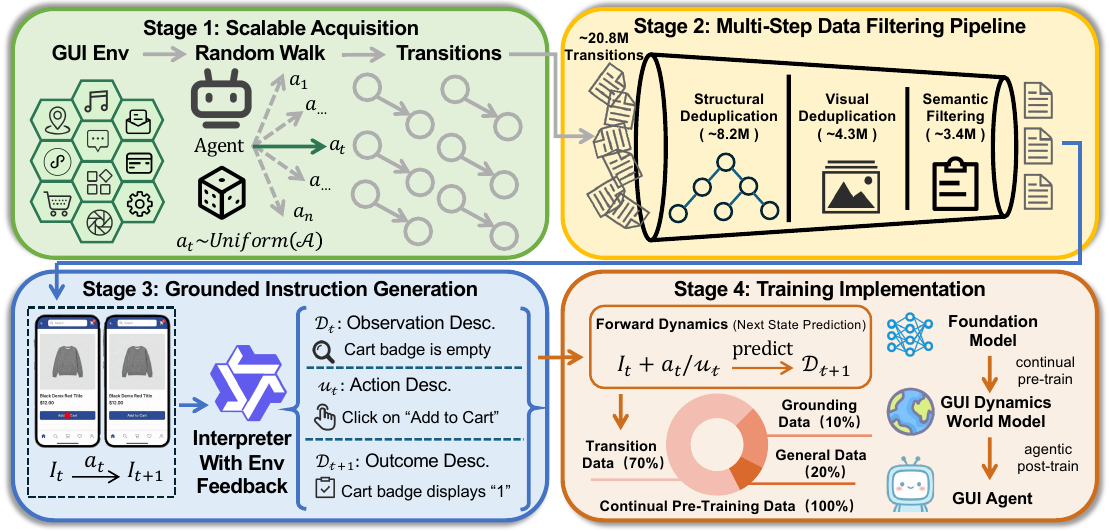}
    \caption{Overview of the proposed \toolname{} framework. 
    \toolname{} consists of four sequential stages: (1) \textbf{Scalable Acquisition}, which autonomously explores diverse GUI applications to generate large-scale raw interaction trajectories; (2) \textbf{Multi-Step Data Filtering Pipeline}, which systematically filters and deduplicates raw interactions based on structural, visual, and semantic criteria; (3) \textbf{Grounded Instruction Generation}, which synthesizes multimodal instructions by interpreting transitions grounded in actual environmental feedback; and (4) \textbf{Training Implementation}, which employs forward dynamics for continual pre-training of the world model, followed by agentic post-training to finalize the GUI agent.}
    \label{fig:pipeline}
\end{figure*}

We conduct extensive empirical evaluations within the WeChat mini-program ecosystem~\cite{WeChatMiniProgramDocs}, one of the world's largest and most diverse digital platforms.
Hosting millions of distinct mini-programs for over a billion monthly active users (MAU), this ecosystem presents a virtually infinite space of interaction logic and visual layouts, demanding that agents master universal interaction dynamics rather than memorizing specific app interfaces.
Notably, \toolname{} demonstrates robust scalability on the WeChat mini-program offline benchmark, achieving an average improvement of +7\% across seven different VLM backbones compared to strong baselines with identical Supervised Fine-Tuning.
Crucially, these benefits are amplified in online navigation: \toolname{} boosts the real-world success rate by 21.9\% during the cold-start phase and maintains an average 15\% lead even after both step-level and multi-step GRPO~\cite{shao2024deepseekmath} alignment.
Furthermore, our experiments yield three critical insights:
(1)~\textbf{Data scaling}: Performance scales log-linearly with synthetic data volume, showing no saturation up to 32B parameters and 3.2B tokens, confirming the scalability of atomic dynamics supervision.
(2)~\textbf{Mechanism}: Forward dynamics proves superior to inverse dynamics, as its higher predictive difficulty compels the model to learn more robust, semantic visual representations.
(3)~\textbf{Generalization}: The learned world model captures universal GUI physics effectively, enabling robust generalization to unseen mini-programs and Android environments, as well as compositional reasoning on complex multi-step tasks never seen during training.

We highlight main contributions as below:
\begin{itemize}
\item We pioneer a shift in GUI agent learning by separating the acquisition of atomic interaction physics from high-level instruction following. To the best of our knowledge, this is the first work to systematically scale environment dynamics as a self-supervised CPT objective for GUI agents, fundamentally bypassing the data scalability limitations of trajectory-based behavioral cloning.
\item We propose \textbf{\toolname{}}, a scalable framework designed to mine training supervision directly from intrinsic environmental signals. 
\item We conduct extensive evaluations validating the performance scaling of \toolname{}. Our results demonstrate robust generalization across both domains and horizons, while ablation studies identify forward dynamics as the optimal objective for GUI world models.
\end{itemize}
\section{Background and Related Work}
\label{sec:related}

Generalist GUI agents autonomously execute user instructions by interacting with Graphical User Interfaces~(GUIs), thereby significantly boosting user productivity in daily tasks~\cite{wang2024gui}.
Pretrained from VLMs, generalist GUI agents are post-trained using Behavioral Cloning~(BC)~\cite{wu2024atlas, lin2025showui, lu2024gui, chen2024guicourse} or multi-step Reinforcement Learning~(RL)~\cite{ye2025mobile, xu2025mobilerlonlineagenticreinforcement, wang2025ui}. 
However, post-training methods struggle under the capability ceiling of the base model~\cite{zhang2025interplaypretrainingmidtrainingrl, yue2025does}, underscoring the necessity of Continual Pre-Training~(CPT) prior to policy alignment.
A fundamental challenge for CPT is the data scalability bottleneck, where collecting high-quality human demonstrations is prohibitively expensive.
To bypass the challenge, existing research mainly uses the following two learning tasks in CPT: (1)~static UI understanding task such as element grounding or screen captioning~\cite{tang2025magicgui, wu2024atlas}, which requires powerful foundation models to label GUI screenshots and lacks grounding in real execution; (2)~dynamic interacting with environments under instructions~\cite{lin2025gui,sun2025genesis,ramrakhya2025scaling,pahuja2025explorer}, which is notoriously difficult due to the low yield rates for high-quality long-horizon samples.
Our \toolname{} proposes a new CPT task: learning on forward dynamics of atomic transitions. 
Such dynamic learning task enables \toolname{} to overcome the bias of distillation caused by lack of grounding, and the fragility of long-horizon trajectory synthesis.

Conceptualizing the GUI as a dynamic system aligns with the broader literature on world models. 
In the specific GUI domain, prior works have explored this direction through visual state prediction~\cite{luo2025vimo,wang2025vagen} and action prediction from state changes~\cite{gao2025uishift}.
However, there still lacks a scalable, task-agnostic CPT objective to master global environment dynamics.
We distinguish our work by leveraging autonomous exploration and intrinsic environmental feedback to internalize a GUI world model of the environment via CPT, establishing a robust dynamics foundation before agentic post-training.

Appendix~\ref{app:related} discusses related work in more detail.

\section{Problem Formulation}
\label{sec:problem}

We formalize the GUI navigation problem not merely as trajectory execution, but as learning the underlying transition dynamics of the interface.

\paragraph{GUI as a state transition graph.}
We model the GUI environment as a directed state transition graph $\mathcal{G} = (\mathcal{S}, \mathcal{E})$. Each node $s \in \mathcal{S}$ represents a unique UI state, formally defined as a tuple $s = (I, X)$, where $I \in \mathbb{R}^{H \times W \times 3}$ is the pixel-level screenshot and $X$ denotes structural metadata (e.g., the Accessibility Tree).
A directed edge $e \in \mathcal{E}$ represents an atomic transition $(s_t, a_t, s_{t+1})$. 
Here, an action $a_t \in \mathcal{A}$ is defined as a tuple $a_t = (\tau, \theta)$, where $\tau$ denotes the operation type (e.g., \textit{click}, \textit{scroll}, \textit{input}) and $\theta$ represents the corresponding parameters (e.g., screen coordinates or input text content), with the full specification of $\mathcal{A}$ detailed in Appendix~\ref{app:action_space}.
Executing $a_t$ on state $s_t$ leads to state $s_{t+1}$.

\paragraph{Forward dynamics (outcome prediction).}
This objective constitutes the agent's world model, predicting the next state $s_{t+1}$ given the current state $s_t$ and action $a_t$. We formulate this by minimizing the negative log-likelihood:
\begin{equation}
    \mathcal{L}_{\text{fwd}} = - \mathbb{E}_{(s_t, a_t, s_{t+1}) \sim \mathcal{D}} [\log P_{\theta}(s_{t+1} \mid s_t, a_t)]
\end{equation}

\paragraph{Inverse dynamics (action inference).}
Inverse dynamics aims to infer the action $a_t$ responsible for the transition between $s_t$ and $s_{t+1}$. The loss function is defined as:
\begin{equation}
    \mathcal{L}_{\text{inv}} = - \mathbb{E}_{(s_t, a_t, s_{t+1}) \sim \mathcal{D}} [\log P_{\theta}(a_t \mid s_t, s_{t+1})]
\end{equation}

\paragraph{Backward dynamics (precondition inference).}
Complementary to forward dynamics, this objective learns to infer the precondition state $s_t$ given the executed action $a_t$ and the outcome $s_{t+1}$. We minimize:
\begin{equation}
    \mathcal{L}_{\text{bwd}} = - \mathbb{E}_{(s_t, a_t, s_{t+1}) \sim \mathcal{D}} [\log P_{\theta}(s_t \mid s_{t+1}, a_t)]
\end{equation}
\section{The \toolname{} Framework}
\label{sec:method}

To instantiate the theoretical world model defined in Section~\ref{sec:problem}, we introduce \toolname{}, a unified framework for learning GUI dynamics from large-scale autonomous interaction. As illustrated in Figure~\ref{fig:pipeline}, \toolname{} consists of four sequential stages: scalable acquisition via autonomous exploration, strict multi-step data filtering pipeline, grounded instruction generation based on environmental feedback, and continual pre-training that optimizes a GUI dynamics world model on the constructed supervision.

\subsection{Scalable Acquisition via Autonomous Exploration}
\label{subsec:acquisition}

We autonomously explore the WeChat mini-program ecosystem to harvest open-domain transitions. Using a distributed fleet of 50 nodes, we employ a structured random walk policy: agents interact with UI elements identified via accessibility trees without human intervention. For each step $t$, we record the transition tuple $(s_t, a_t, s_{t+1})$, where $s_t = (I_t, X_t)$ denotes the screenshot and structural metadata. This process yielded over 20.8M raw transitions over one month.

\subsection{Multi-Step Data Filtering Pipeline}
\label{sec:filtering}

We refine $\mathcal{D}_{\text{raw}}$ into a high-quality corpus $\mathcal{D}_{\text{cpt}}$ via a three-stage pipeline(details in Appendix~\ref{app:data_filtering}):
(1) \textbf{Structural Deduplication:} MinHash signatures on accessibility trees remove repeated content templates, reducing data to 8.2M.
(2) \textbf{Visual Deduplication:} pixel-level pHash/dHash identify visually static transitions (e.g., buffering or invisible DOM updates), narrowing it to 4.3M.
(3) \textbf{Semantic Flitering:} A VLM filters system errors and rendering artifacts by verifying action-feedback consistency, resulting in a final set of 3.4M transitions.

\subsection{Grounded Instruction Generation}
\label{subsec:synthesis}

Following the data filtering stage, we convert the cleaned transition tuples into high-quality multimodal instructions using a vision-language model (Qwen3-VL-235B-A22B-Instruct~\cite{Qwen3-VL}) as a semantic synthesizer. Unlike standard distillation methods, our approach utilizes the actual next state $I_{t+1}$ provided directly by the environment, with the synthesizer acting solely as an interpreter to articulate the observed transition. This strategy effectively prevents the synthesizer from generating hallucinations typically found in purely model-based approaches.

For each transition $(s_t, a_t, s_{t+1})$, we construct a visual prompt that includes the pre-transition screenshot $I_t$, annotated with a visual marker $m(a_t)$ indicating the action location (e.g., a circle highlighting the clicked area), along with the post-transition screenshot $I_{t+1}$. Given this prompt, the synthesizer generates a structured output consisting of three distinct components:
(1) \textbf{Observation description ($\mathcal{D}_t$):} a concise description of the initial GUI state before the action.
(2) \textbf{Action description ($u_t$):} a brief natural-language summary describing the executed action.
(3) \textbf{Outcome description ($\mathcal{D}_{t+1}$):} a precise description of the resulting GUI state after the action.

By generating instructions directly grounded in actual environmental state transitions, this stage produces large-scale, reliable, and high-fidelity data for CPT.

\subsection{Training Implementation}
\label{subsec:training}

\subsubsection{Continual Pretraining}

Using the curated dataset $\mathcal{D}_{\text{cpt}}$ and generated grounded instructions, we construct a unified training corpus designed to facilitate understanding of GUI dynamics.

We define three distinct categories of training tasks to comprehensively capture GUI dynamics:

\textbf{Forward dynamics:} Given the screenshot of an initial state and an action (either as an exact action or a natural-language summary), the model predicts the resulting GUI state description.  
\textit{Input}: $\mathcal{V} = [I_t, a_t]$ or $\mathcal{V} = [I_t, u_t]$; \textit{Target}: $\mathbf{y} = \mathcal{D}_{t+1}$.

\textbf{Inverse dynamics:} Given screenshots of the initial and resulting states, the model predicts the action connecting them.  
\textit{Input}: $\mathcal{V} = [I_t, I_{t+1}]$; \textit{Target}: $\mathbf{y} = u_t$ or $\mathbf{y} = a_t$.
    
\textbf{Backward dynamics:} Given the screenshot of a resulting state and an action summary, the model predicts a description of the initial GUI state.  
\textit{Input}: $\mathcal{V} = [u_t, I_{t+1}]$; \textit{Target}: $\mathbf{y} = \mathcal{D}_{t}$.

\textbf{Data Mixing.}
We construct the final training set $\mathcal{D}_{\text{mix}}$ by strategically combining three distinct data sources:
(1) \textbf{GUI dynamics (70\%):} Informed by our ablation study (Section~\ref{subsec:ablation}), we exclusively utilize Forward Dynamics ($I_t, a_t \to \mathcal{D}_{t+1}$) to maximize generative world modeling capabilities;
(2) \textbf{General multimodal data (20\%):} General data~\cite{zhang2025bee} acts as a regularizer to prevent the catastrophic forgetting of foundational capabilities;
(3) \textbf{UI grounding data (10\%):} Grounding data is incorporated to preserve fine-grained spatial localization skills, ensuring that precise execution capabilities are not compromised.
We detail the specific training hyperparameters in Appendix~\ref{app:cpt_hyperparams}.

\subsubsection{Agentic Post-training.}
Following the CPT phase, we conduct an agentic post-training stage. In this phase, the model is fine-tuned on high-quality GUI agent trajectories to explicitly align the GUI world model with user intents, bridging the gap between physical understanding and task execution. We detail the specific training hyperparameters in Appendix~\ref{app:sft_hyperparams}.
\section{Experiments}
\label{sec:experiments}

In this section, we systematically evaluate the \toolname{} framework.
Our evaluation is structured to answer four critical questions:
(1) \textbf{Scalability:} Can autonomous exploration serve as a scalable data source that yields consistent performance gains? (Section~\ref{subsec:scaling})
(2) \textbf{Online efficacy:} Does the learned world model translate effectively to real-world online navigation and act as a persistent prior that synergizes with agentic post-training? (Section~\ref{subsec:online_rl})
(3) \textbf{Mechanism:} Which self-supervised objective yields the most significant performance gains for downstream GUI navigation? (Section~\ref{subsec:ablation})
(4) \textbf{Generalization:} Does the learned world model generalize to unseen apps, novel domains (Android), and complex compositional tasks? (Section~\ref{subsec:generalization})

\subsection{Scaling Laws of \toolname{}}
\label{subsec:scaling}

\begin{table*}[t]
\centering
\renewcommand{\arraystretch}{1.15}
\setlength{\tabcolsep}{3.5pt} 
\caption{\textbf{Scaling Laws of \toolname{}.} Evaluation of Exact Match (EM) scores on the held-out offline benchmark. The ``0\% (SFT)'' column represents the standard Supervised Fine-Tuning baseline. By scaling the volume of continual pre-training dynamics data from 12.5\% to 100\%, \toolname{} achieves consistent performance gains across all seven model backbones. The results validate a clear positive correlation between dynamics data volume and downstream navigation capability.}
\label{tab:scaling_results}

\begin{tabular}{l c cc ccccc}
\toprule
& & \multicolumn{2}{c}{\textbf{Reference}} & \multicolumn{5}{c}{\textbf{\toolname{} Data Scale (+ SFT)}} \\
\cmidrule(lr){3-4} \cmidrule(lr){5-9} 

\textbf{Model} & \textbf{Size} & \textbf{Base (Zero-shot)} & \textbf{w/ grounding} & \textbf{0\% (SFT)} & \textbf{12.5\%} & \textbf{25\%} & \textbf{50\%} & \textbf{100\% (Ours)} \\
\midrule

\multicolumn{8}{l}{\textit{\textbf{I. Qwen3-VL Series with \toolname{}}}} \\
\cmidrule(r){1-3}
Qwen3-VL & 2B & 2.4 & 36.5 & 47.9 & 49.4 & 51.3 & 51.7 & 52.7 \\
Qwen3-VL & 4B & 41.7 & 52.0 & 57.3 & 58.3 & 59.0 & 59.3 & 60.0 \\
Qwen3-VL & 8B & 42.3 & 54.9 & 58.1 & 58.0 & 59.6 & 60.4 & 60.8 \\
Qwen3-VL & 32B & 47.1 & 61.2 & 61.5 & 61.0 & 63.7 & 62.5 & 65.0 \\
\midrule

\multicolumn{8}{l}{\textit{\textbf{II. Qwen2.5-VL Series with \toolname{}}}} \\
\cmidrule(r){1-3}
Qwen2.5-VL & 3B &  1.1 & 27.9 & 47.7 & 49.2 & 48.5 & 49.6 & 50.6 \\
Qwen2.5-VL & 7B &  13.0 & 37.8 & 51.0 & 53.5 & 53.3 & 53.4 & 55.7 \\
Qwen2.5-VL & 32B & 33.9 & 56.5 & 60.5 & 61.6 & 63.2 & 64.2 & 64.8 \\
\bottomrule
\end{tabular}
\end{table*}

A central premise of our framework is that autonomous environmental feedback can overcome the data scarcity bottleneck. We first investigate the scaling properties of \toolname{}.

\subsubsection{Setup}

\textbf{Training configurations.} During the CPT phase, we utilize the data mixture detailed in Section~\ref{subsec:training}, totaling approximately 5M samples (3.2B tokens). For the subsequent agentic post-training phase, we employ 8K high-quality GUI navigation samples. To verify data scaling laws, we vary the volume of CPT data injected from 0\% to 100\%.
\textbf{Evaluation Protocol.} To support large-scale experiments, we evaluate performance on a held-out offline benchmark comprising 8K diverse mini-program tasks not seen during training. Following prior works~\cite{wu2024atlas, zhang2024android, li2024effects}, we report Exact Match (EM), which requires both the action type and its parameters to be correct. We assess performance across Qwen3-VL~\cite{Qwen3-VL} (2B, 4B, 8B, 32B) and Qwen2.5-VL~\cite{Qwen2.5-VL} (3B, 7B, 32B) families.
We compare our method against two distinct baselines:
(1) \textbf{Base Models:} We evaluate raw base models in a zero-shot setting. To decouple planning from grounding, we also report a ``w/ grounding'' setting, where an external specialized grounding model based on Qwen2.5-VL-32B is employed to predict an action based on the natural language action descriptions;
(2) \textbf{SFT-only Baselines:} Models fine-tuned solely on the post-training data without any dynamics CPT (denoted as 0\% data scale).

\begin{figure}
    \centering
    \begin{subfigure}[t]{0.48\linewidth}
        \centering
        \includegraphics[width=\linewidth]{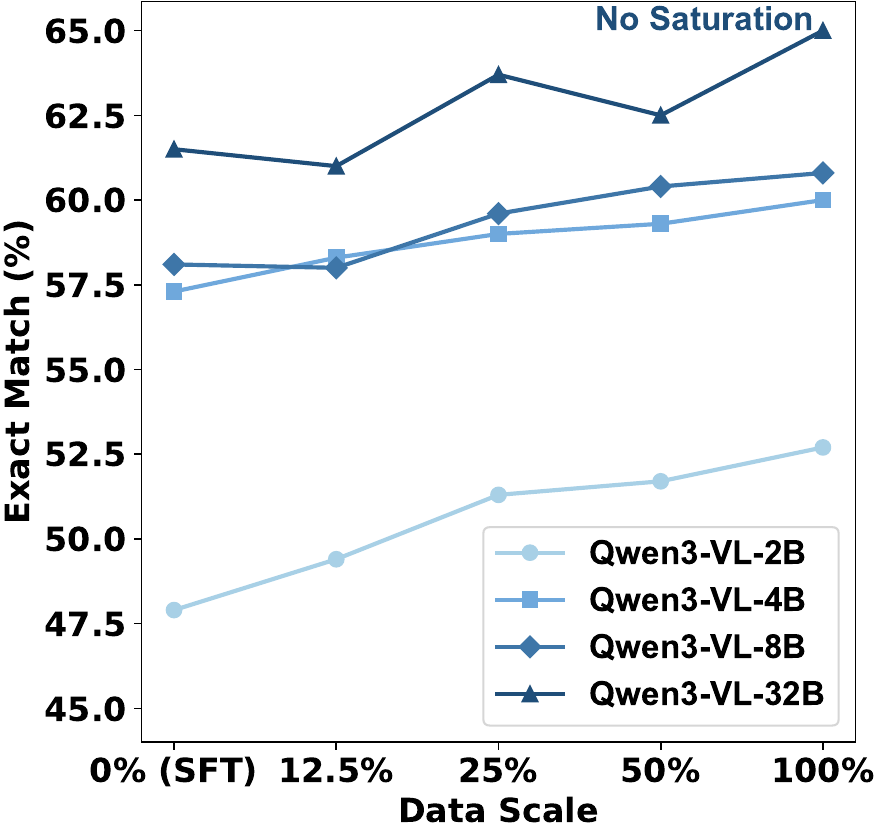}
        \caption{Data scaling}
        \label{fig:scaling-a}
    \end{subfigure}
    \hfill
    \begin{subfigure}[t]{0.45\linewidth}
        \centering
        \includegraphics[width=\linewidth]{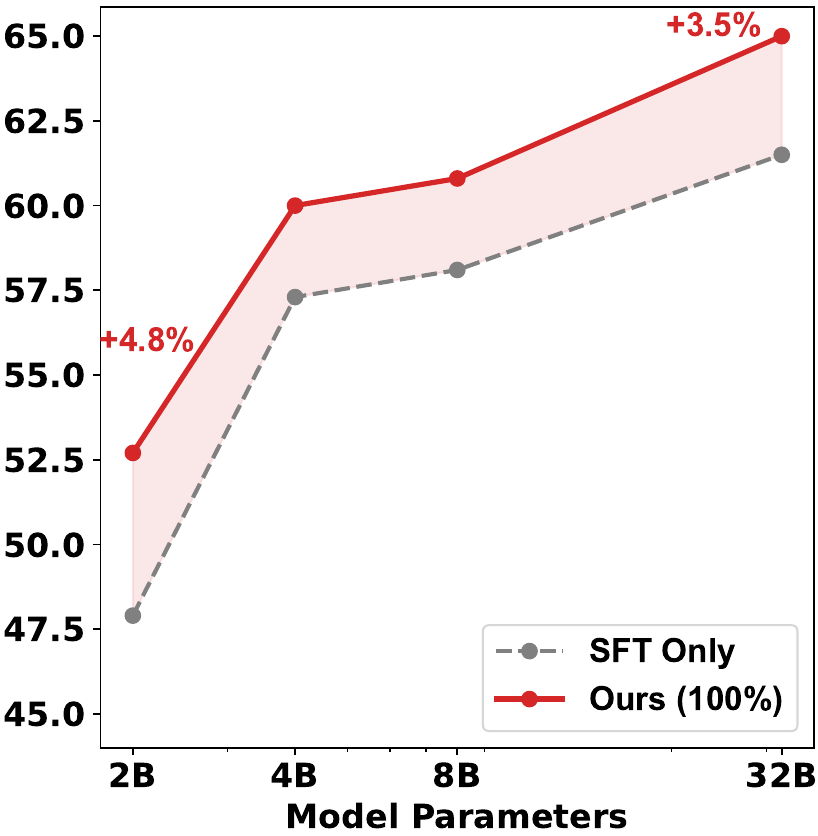}
        \caption{Model scaling}
        \label{fig:scaling-b}
    \end{subfigure}
    \caption{Scaling behavior of Qwen3-VL series models.}
    \label{fig:scaling}
\end{figure}

\subsubsection{Results}

As presented in Table~\ref{tab:scaling_results} and visualized in Figure~\ref{fig:scaling}, our empirical findings validate the existence of scaling laws for GUI world models across both data and model dimensions.

\paragraph{Data scaling.}
We observe a consistent positive correlation between the volume of pre-training dynamics data and downstream performance.
Starting from the standard SFT baseline (0\% data scale), the injection of our dynamics data yields consistent improvements across all model sizes.
For instance, the Qwen3-VL-2B model improves from 47.9\% (0\%) to 52.7\% (100\%), while the larger Qwen3-VL-32B advances from 61.5\% to 65.0\%.
The performance curves exhibit a strong log-linear trend, confirming that autonomous exploration serves as a scalable data source.

\paragraph{Model scaling.}
\toolname{} demonstrates robust scalability with model size.
Comparing the Qwen3-VL series, the EM score improves steadily from 52.7\% (2B) to 65.0\% (32B) as parameters increase.
Crucially, the performance gains over the SFT baseline remain robust across all scales.
This indicates that the internalized interaction dynamics provide fundamental physical knowledge that complements even larger capacity models, rather than diminishing as model size grows.

\paragraph{Absence of saturation.}
Notably, we observe no signs of performance saturation across either dimension.
Even at the maximal data scale(5M samples, 3.2B tokens) and model size (32B), the trend lines maintain their upward trajectory.
This suggests that further scaling would likely yield continued gains.

\subsection{Online Evaluation: Synergy with Post-Training}
\label{subsec:online_rl}

\begin{table}[t]
\centering
\small
\renewcommand{\arraystretch}{1.25}
\setlength{\tabcolsep}{3pt} 

\caption{\textbf{Online Evaluation across Training Stages.} Success Rate (SR) on live mini-programs. \textbf{$\Delta$} shows the gain from \toolname{}.}
\label{tab:online_rl}

\begin{tabular*}{\columnwidth}{l @{\extracolsep{\fill}} c c c}
\toprule
\multirow{2}{*}{\textbf{Training Stage}} & \multicolumn{2}{c}{\textbf{SR (\%)}} & \multirow{2}{*}{\textbf{$\Delta$}} \\
\cmidrule(lr){2-3}
& \textbf{w/o CPT} & \textbf{\toolname{}} & \\
\midrule

\multicolumn{4}{l}{\textit{\textbf{Qwen3-VL-8B}}} \\
\hspace{1em} SFT + Step-level GRPO 
 & 27.5 & \textbf{30.9} & \textcolor{green!60!black}{+12.4\%} \\

\midrule

\multicolumn{4}{l}{\textit{\textbf{Qwen2.5-VL-32B}}} \\
\hspace{1em} SFT (Cold Start) 
 & 24.2 & \textbf{29.5} & \textcolor{green!60!black}{+21.9\%} \\
 
\hspace{1em} + Step-level GRPO 
 & 32.2 & \textbf{38.9} & \textcolor{green!60!black}{+20.8\%} \\
 
\hspace{1em} + Multi-step GRPO 
 & 39.6 & \textbf{44.3} & \textcolor{green!60!black}{+11.9\%} \\

\bottomrule
\end{tabular*}
\end{table}

While offline metrics quantify predictive accuracy, the ultimate test lies in online execution. We investigate the effectiveness of \toolname{} when integrated into a progressive post-training pipeline.

\subsubsection{Setup}

\textbf{Training configurations.} We conduct a comparative analysis between agents initialized with and without our proposed CPT (\toolname{} vs. w/o CPT). The post-training pipeline consists of three progressive stages:
(1) \textbf{SFT (cold start)}: Supervised fine-tuning using simple samples from the SFT dataset to establish basic instruction-following capabilities;
(2) \textbf{Step-level GRPO}: Optimization on hard samples from the SFT dataset using dense step-wise rewards to refine complex reasoning;
(3) \textbf{Multi-step GRPO}: Further aligning using approximately 1K instructions in the real environment, utilizing a VLM as an outcome reward model to provide sparse trajectory-level reward.

\textbf{Evaluation protocol.} We report the Success Rate (SR) on a held-out benchmark of 149 instructions covering diverse mini-programs. All evaluation episodes are executed in the real online environment, and task success is manually verified by human annotators to ensure rigorous correctness.

\subsubsection{Results}
As shown in Table~\ref{tab:online_rl}, \toolname{} provides a robust performance foundation that synergizes effectively with the entire post-training pipeline.

\paragraph{Consistent superiority across post-training stages.}
\toolname{} establishes an immediate advantage in the initial SFT phase and, crucially, maintains this lead throughout the subsequent RL alignment steps.
Across both step-level and multi-step GRPO stages, agents initialized with our world model consistently outperform the baselines (e.g., a +20.8\% relative gain in step-level GRPO).
This persistence demonstrates that the internalized physical understanding is not overwritten by optimization but rather serves as an intrinsic prior that facilitates more efficient policy learning.

\paragraph{Amplified benefits in online environments.}
Notably, the performance gains yielded by \toolname{} are more pronounced in the online setting compared to the offline benchmark (Section~\ref{subsec:scaling}).
While offline evaluation relies on static matching, online execution demands resilience to dynamic rendering variations and error recovery.
The amplified gain in this challenging environment highlights that our world model equips the agent with the robust physical understanding necessary for real-world interaction.

\subsection{Mechanism: The Superiority of Forward Dynamics}
\label{subsec:ablation}

\begin{table*}[t]
\centering
\footnotesize
\setlength{\tabcolsep}{5pt}
\caption{\textbf{Ablation Study on Continual Pre-training Objectives.} We compare different dynamics tasks against the non-CPT baseline. \textbf{Key Insight:} Forward dynamics ($I_t, u_t/a_t \to \mathcal{D}_{t+1}$) consistently achieve the highest performance across all scales. The ``Overall'' column reports the average TM and EM scores across the three model scales. Best results are bolded, and second-best are underlined.}
\label{tab:ablation_reordered}

\begin{tabular}{llcccccccc}
\toprule
\multirow{2}{*}{\textbf{Dynamics}} & \multirow{2}{*}{\textbf{Formulation}} & \multicolumn{2}{c}{\textbf{Qwen3-VL-2B}} & \multicolumn{2}{c}{\textbf{Qwen3-VL-4B}} & \multicolumn{2}{c}{\textbf{Qwen2.5-VL-3B}} & \multicolumn{2}{c}{\textbf{Overall}} \\
\cmidrule(lr){3-4} \cmidrule(lr){5-6} \cmidrule(lr){7-8} \cmidrule(lr){9-10}
& & \textbf{TM} & \textbf{EM} & \textbf{TM} & \textbf{EM} & \textbf{TM} & \textbf{EM} & \textbf{TM} & \textbf{EM} \\
\midrule

\multicolumn{2}{l}{\textit{Baseline (w/o CPT)}} & 75.9 & 47.9 & 79.2 & 57.3 & 75.7 & 47.7 & 76.9 & 51.0 \\

\midrule

\multirow{2}{*}{Forward} & $I_t, u_t \to \mathcal{D}_{t+1}$            & \textbf{77.7} & \textbf{52.7} & \textbf{81.0} & \textbf{60.3} & \underline{76.7} & 48.8 & \textbf{78.5} & \underline{53.9} \\
                         & $I_t, a_t \to \mathcal{D}_{t+1}$            & \underline{77.6} & \textbf{52.7} & \underline{80.3} & \underline{60.0} & \textbf{76.9} & \textbf{50.6} & \underline{78.3} & \textbf{54.4} \\

\midrule

\multirow{4}{*}{Inverse} & $I_t, I_{t+1} \to u_t$                      & 75.8 & 49.5 & 75.8 & 49.5 & 75.3 & \underline{49.0} & 75.6 & 49.3 \\
                         & $I_t, I_{t+1} \to a_t$                      & 75.6 & 46.4 & 77.8 & 50.9 & 75.6 & 47.2 & 76.3 & 48.2 \\
\cmidrule(lr){2-10}
                         & $I_t, \mathcal{D}_{t+1} \to u_t$            & 76.3 & 50.2 & 79.0 & 57.4 & 76.2 & 48.3 & 77.2 & 52.0 \\
                         & $I_t, \mathcal{D}_{t+1} \to a_t$            & 75.7 & 46.6 & 78.0 & 52.5 & 75.8 & 47.6 & 76.5 & 48.9 \\
\midrule

Backward                 & $u_t, I_{t+1} \to \mathcal{D}_t$            & 77.1 & \underline{50.4} & 80.0 & 58.7 & 76.3 & 47.5 & 77.8 & 52.2 \\
\bottomrule
\end{tabular}
\end{table*}

We dissect the contributions of different CPT objectives to understand the source of the performance gains.

\subsubsection{Setup}

\textbf{Training configuration.} To isolate the impact of objective formulation, we conduct controlled experiments across three model scales (Qwen3-VL-2B, 4B, and Qwen2.5-VL-3B). We compare the three distinct categories of self-supervised tasks detailed in Section~\ref{subsec:training}, while keeping the underlying UI transitions corpus constant (5M samples).

\textbf{Evaluation protocol.} We adhere to the identical SFT and evaluation dataset settings established in Section~\ref{subsec:scaling}. In addition to the previously defined EM, we report Type Match (TM), which solely evaluates the classification accuracy of the predicted action type against the ground truth.

\subsubsection{Results and Analysis}

As detailed in Table~\ref{tab:ablation_reordered}, our ablation study reveals critical insights into how different physical objectives shape the agent's representations.

\paragraph{The dominance of forward dynamics.}
Forward dynamics dominates, with the simplest formulation $I_t, a_t \to \mathcal{D}_{t+1}$ objective achieving the highest EM (54.4\%).
Predicting \textit{``what happens next''} forces the model to encode the causal logic of the environment.
Crucially, distinct from methods employing separate world models for planning~\cite{luo2025vimo,li2025mobileworldbench}, our approach embeds this transition reasoning directly into the agent's representation. This eliminates the need for external models, allowing the agent to reason about dynamics implicitly.

\paragraph{The failure of inverse dynamics.}
A counter-intuitive finding is the poor performance of Inverse Dynamics. Despite sharing the identical action space to the navigation task, variants like $I_t, I_{t+1} \to a_t$ yield an Overall EM of only 48.2\%, performing even worse than the non-CPT baseline (51.0\%). We attribute this degradation to two factors:
\begin{itemize}
    \item \textbf{Sparse supervision signal:} Unlike forward dynamics which generates rich textual descriptions, the inverse task only outputs action types and coordinates. This limited token coverage and lack of diversity restrict the model's ability to learn dense, semantically meaningful representations during CPT.
    \item \textbf{Insufficient task difficulty:} As illustrated in Figure~\ref{fig:loss_curve}, inverse dynamics converges rapidly to a low loss value.
This indicates that the task lacks sufficient predictive complexity.
Upon rapid convergence, the objective ceases to provide a meaningful gradient signal, rendering it ill-suited for effective CPT where sustained predictive challenge is essential~\cite{velasco2025rethinking, agrawal2023corpus, marion2023less, geirhos2020shortcut}.
\end{itemize}

\paragraph{Backward dynamics and temporal directionality.}
While backward dynamics ($u_t, I_{t+1}\to \mathcal{D}_t$) also requires learning environment physics, it lags behind forward dynamics (52.2\% vs. 54.4\%).
This indicates that while ``reconstructing the past'' offers some regularization benefits, it is less effective than ``predicting the future'', which directly aligns with the agent's inferential goal of planning subsequent steps based on current observations.

\subsection{Cross-Domain and Compositional Generalization of World Model}
\label{subsec:generalization}

\begin{table}[t]
\centering
\scriptsize
\setlength{\tabcolsep}{3.5pt}
\caption{\textbf{Cross-Domain and Compositional Generalization.} We evaluate the learned world model across varying distribution shifts (mini-programs $\to$ Android) and reasoning horizons (atomic L1 $\to$ compositional L2). \toolname{} demonstrates robust cross-domain transfer and the ability to chain atomic rules for multi-step reasoning. We abbreviate mini-programs as mini-P.}
\label{tab:generalization_3models}

\begin{tabular}{lllccc}
\toprule
\multirow{2}{*}{\textbf{Lvl}} & \multirow{2}{*}{\textbf{Task}} & \multirow{2}{*}{\textbf{Model}} & \textbf{Seen} & \textbf{Unseen} & \textbf{OOD} \\
 & & & \tiny{\textbf{mini-P.}} & \tiny{\textbf{mini-P.}} & \tiny{\textbf{Android}} \\
\midrule

\multirow{6}{*}{\textbf{L1}}
 & \multirow{3}{*}{\shortstack[l]{Forward \\ \tiny{$I_t, u_t \to \mathcal{D}_{t+1}$}}}
   & Qwen2.5-VL-7B & 41.2 & 43.1 & 46.5 \\
 & & Qwen2.5-VL-32B & 48.5 & 49.6 & \textbf{65.8} \\
 & & \cellcolor{gray!10}\textbf{\toolname{}-7B-F} & \cellcolor{gray!10}\textbf{53.1} & \cellcolor{gray!10}\textbf{56.9} & \cellcolor{gray!10}49.1 \\
 \cmidrule{2-6}
 & \multirow{3}{*}{\shortstack[l]{Inverse \\ \tiny{$I_t, I_{t+1} \to u_t$}}}
   & Qwen2.5-VL-7B & 17.0 & 18.4 & 31.1 \\
 & & Qwen2.5-VL-32B & 37.0 & 37.5 & \textbf{67.3} \\
 & & \cellcolor{gray!10}\textbf{\toolname{}-7B-I} & \cellcolor{gray!10}\textbf{62.3} & \cellcolor{gray!10}\textbf{58.9} & \cellcolor{gray!10}55.3 \\
\midrule

\multirow{6}{*}{\textbf{L2}}
 & \multirow{3}{*}{\shortstack[l]{Forward \\ \tiny{$I_t, u_t, u_{t+1} \to \mathcal{D}_{t+2}$}}}
   & Qwen2.5-VL-7B & 41.3 & 41.1 & / \\
 & & Qwen2.5-VL-32B & 46.9 & 46.3 & / \\
 & & \cellcolor{gray!10}\textbf{\toolname{}-7B-F} & \cellcolor{gray!10}\textbf{47.0} & \cellcolor{gray!10}\textbf{47.2} & \cellcolor{gray!10}/ \\
 \cmidrule{2-6}
 & \multirow{3}{*}{\shortstack[l]{Inverse \\ \tiny{$I_t, I_{t+2} \to u_t$}}}
   & Qwen2.5-VL-7B & 6.7 & 6.2 & / \\
 & & Qwen2.5-VL-32B & 26.5 & 26.1 & / \\
 & & \cellcolor{gray!10}\textbf{\toolname{}-7B-I} & \cellcolor{gray!10}\textbf{38.7} & \cellcolor{gray!10}\textbf{37.6} & \cellcolor{gray!10}/ \\
\bottomrule
\end{tabular}
\end{table}

Finally, we investigate whether the agent has merely memorized specific transition pairs or internalized a robust world model capable of cross-domain transfer and compositional reasoning.
To strictly assess this, we define two levels of tasks:
Level 1 (L1, atomic) utilizes standard single-step transitions ($t \to t+1$) to evaluate cross-domain transfer performance;
Level 2 (L2, compositional) introduces multi-step chaining ($t \to t+2$) which never seen during training to assess compositional generalization, testing the ability to combine atomic physical rules for long-horizon prediction.

\subsubsection{Setup}

\textbf{Dataset construction.} We construct a controlled evaluation benchmark using atomic transitions from mini-programs, partitioned at the application level into seen and unseen groups. We randomly sample 1.2M transitions for training, while reserving 10K samples for L1 test sets and constructing 10K samples for L2 test sets within each group. To evaluate out-of-distribution (OOD) robustness, we additionally sample 10K random transitions from AndroidControl~\cite{li2024effects}.

\textbf{Training configurations.} We train two specialized variants of \toolname{}-7B using the same data mixture as in Section~\ref{subsec:ablation}. Specifically, \toolname{}-7B-F is trained on L1 forward tasks ($I_t, u_t \to \mathcal{D}_{t+1}$), and \toolname{}-7B-I is trained on L1 inverse tasks ($I_t, I_{t+1} \to u_t$). Both models are trained on approximately 1.7M total samples.

\textbf{Evaluation protocol.} We employ VLM-as-a-Judge to handle the open-ended nature of generative predictions.
For forward dynamics, we explicitly query the model to predict 5 distinct UI elements of the outcome page to facilitate objective verification via a VLM judge.
For inverse dynamics, the model infers the first action given the start and end states, with correctness verified by a VLM judge.

\subsubsection{Results}

As shown in Table~\ref{tab:generalization_3models}, \toolname{} demonstrates robustness across both domain shifts and reasoning horizons.

\paragraph{Cross-domain generalization.}
Our model exhibits strong transfer capabilities across distribution shifts.
Notably, on L1 Forward tasks, even under a challenging format shift where the model must adapt from generating holistic descriptions (training) to listing specific elements (inference), \toolname{}-7B still achieves 56.9 on unseen mini-programs, surpassing the significantly larger Base-32B (49.6).
This indicates that our method learns universal interaction logic rather than memorizing app-specific layouts. Furthermore, on the challenging OOD Android benchmark, \toolname{}-7B consistently outperforms its base model across all tasks, confirming effective cross-platform transfer.

\paragraph{Compositional generalization.}
Crucially, although trained strictly on single-step transitions ($s_t \rightarrow s_{t+1}$), \toolname{} exhibits robust compositional generalization on Level 2 tasks.
This implies that \toolname{} does not merely memorize single-step transitions but enables the agent to mentally ``chain'' atomic physical rules. This capability equips the agent with the look-ahead simulation capabilities necessary for long-horizon planning.
\section{Conclusion}
We introduce \toolname{}, a framework that overcomes the data scalability bottleneck by learning a GUI world model from autonomous exploration.
By shifting from trajectory imitation to mastering atomic interaction physics, we unlock a scalable supervision source independent of human annotation.
Our results identify forward dynamics as the critical objective for this capability, yielding log-linear performance scaling on the massive WeChat ecosystem with no signs of saturation up to 32B parameters and 3.2B tokens.
Crucially, this internalized world model acts as a persistent policy prior: it boosts real-world online performance by over 15\% and serves as a robust foundation for agentic post-training.
By separating the acquisition of interaction physics from the execution of user intent, \toolname{} provides a scalable pathway toward generalist agents capable of navigating complex, open-domain interfaces.

\bibliography{main} 

\begin{thebibliography}{56}
\providecommand{\natexlab}[1]{#1}
\providecommand{\url}[1]{\texttt{#1}}
\expandafter\ifx\csname urlstyle\endcsname\relax
  \providecommand{\doi}[1]{doi: #1}\else
  \providecommand{\doi}{doi: \begingroup \urlstyle{rm}\Url}\fi

\bibitem[Agrawal and Singh(2023)]{agrawal2023corpus}
Ameeta Agrawal and Suresh Singh.
\newblock Corpus complexity matters in pretraining language models.
\newblock In \emph{Proceedings of The Fourth Workshop on Simple and Efficient
  Natural Language Processing (SustaiNLP)}, pages 257--263, 2023.

\bibitem[{Anthropic}(2025{\natexlab{a}})]{anthropic2025claudeopus45}
{Anthropic}.
\newblock Claude opus 4.5, 2025{\natexlab{a}}.
\newblock URL \url{https://www.anthropic.com/claude/opus}.
\newblock Accessed: 2026-01-29.

\bibitem[{Anthropic}(2025{\natexlab{b}})]{anthropic2025claudesonnet45}
{Anthropic}.
\newblock Claude sonnet 4.5, 2025{\natexlab{b}}.
\newblock URL \url{https://www.anthropic.com/news/claude-sonnet-4-5}.
\newblock Accessed: 2026-01-29.

\bibitem[Assran et~al.(2023)Assran, Duval, Misra, Bojanowski, Vincent, Rabbat,
  LeCun, and Ballas]{assran2023self}
Mahmoud Assran, Quentin Duval, Ishan Misra, Piotr Bojanowski, Pascal Vincent,
  Michael Rabbat, Yann LeCun, and Nicolas Ballas.
\newblock Self-supervised learning from images with a joint-embedding
  predictive architecture.
\newblock In \emph{Proceedings of the IEEE/CVF Conference on Computer Vision
  and Pattern Recognition}, pages 15619--15629, 2023.

\bibitem[Bai et~al.(2025{\natexlab{a}})Bai, Cai, Chen, Chen, Chen, Cheng, Deng,
  Ding, Gao, Ge, Ge, Guo, Huang, Huang, Huang, Hui, Jiang, Li, Li, Li, Li, Lin,
  Lin, Liu, Liu, Liu, Liu, Liu, Liu, Lu, Luo, Lv, Men, Meng, Ren, Ren, Song,
  Sun, Tang, Tu, Wan, Wang, Wang, Wang, Wang, Xie, Xu, Xu, Xu, Yang, Yang,
  Yang, Yang, Yu, Zhang, Zhang, Zhang, Zheng, Zhong, Zhou, Zhou, Zhou, Zhu, and
  Zhu]{Qwen3-VL}
Shuai Bai, Yuxuan Cai, Ruizhe Chen, Keqin Chen, Xionghui Chen, Zesen Cheng,
  Lianghao Deng, Wei Ding, Chang Gao, Chunjiang Ge, Wenbin Ge, Zhifang Guo,
  Qidong Huang, Jie Huang, Fei Huang, Binyuan Hui, Shutong Jiang, Zhaohai Li,
  Mingsheng Li, Mei Li, Kaixin Li, Zicheng Lin, Junyang Lin, Xuejing Liu,
  Jiawei Liu, Chenglong Liu, Yang Liu, Dayiheng Liu, Shixuan Liu, Dunjie Lu,
  Ruilin Luo, Chenxu Lv, Rui Men, Lingchen Meng, Xuancheng Ren, Xingzhang Ren,
  Sibo Song, Yuchong Sun, Jun Tang, Jianhong Tu, Jianqiang Wan, Peng Wang,
  Pengfei Wang, Qiuyue Wang, Yuxuan Wang, Tianbao Xie, Yiheng Xu, Haiyang Xu,
  Jin Xu, Zhibo Yang, Mingkun Yang, Jianxin Yang, An~Yang, Bowen Yu, Fei Zhang,
  Hang Zhang, Xi~Zhang, Bo~Zheng, Humen Zhong, Jingren Zhou, Fan Zhou, Jing
  Zhou, Yuanzhi Zhu, and Ke~Zhu.
\newblock Qwen3-vl technical report.
\newblock 2025{\natexlab{a}}.
\newblock URL \url{https://arxiv.org/abs/2511.21631}.

\bibitem[Bai et~al.(2025{\natexlab{b}})Bai, Chen, Liu, Wang, Ge, Song, Dang,
  Wang, Wang, Tang, Zhong, Zhu, Yang, Li, Wan, Wang, Ding, Fu, Xu, Ye, Zhang,
  Xie, Cheng, Zhang, Yang, Xu, and Lin]{Qwen2.5-VL}
Shuai Bai, Keqin Chen, Xuejing Liu, Jialin Wang, Wenbin Ge, Sibo Song, Kai
  Dang, Peng Wang, Shijie Wang, Jun Tang, Humen Zhong, Yuanzhi Zhu, Mingkun
  Yang, Zhaohai Li, Jianqiang Wan, Pengfei Wang, Wei Ding, Zheren Fu, Yiheng
  Xu, Jiabo Ye, Xi~Zhang, Tianbao Xie, Zesen Cheng, Hang Zhang, Zhibo Yang,
  Haiyang Xu, and Junyang Lin.
\newblock Qwen2.5-vl technical report.
\newblock \emph{arXiv preprint arXiv:2502.13923}, 2025{\natexlab{b}}.

\bibitem[ByteDance-Seed(2025)]{Seed1.8}
ByteDance-Seed.
\newblock Seed1.8 model card: Towards generalized real-world agency, 2025.
\newblock Accessed: 2026-01-29.

\bibitem[Chen et~al.(2024)Chen, Cui, Hu, Qin, Fang, Zhao, Wang, Liu, Chen, Huo,
  et~al.]{chen2024guicourse}
Wentong Chen, Junbo Cui, Jinyi Hu, Yujia Qin, Junjie Fang, Yue Zhao, Chongyi
  Wang, Jun Liu, Guirong Chen, Yupeng Huo, et~al.
\newblock Guicourse: From general vision language models to versatile {GUI}
  agents.
\newblock \emph{arXiv preprint arXiv:2406.11317}, 2024.

\bibitem[Cheng et~al.(2024)Cheng, Sun, Chu, Xu, Li, Zhang, and
  Wu]{cheng2024seeclick}
Kanzhi Cheng, Qiushi Sun, Yougang Chu, Fangzhi Xu, Yantao Li, Jianbing Zhang,
  and Zhiyong Wu.
\newblock Seeclick: Harnessing {GUI} grounding for advanced visual {GUI}
  agents.
\newblock \emph{arXiv preprint arXiv:2401.10935}, 2024.

\bibitem[Gao et~al.(2025)Gao, Zhang, and Xu]{gao2025uishift}
Longxi Gao, Li~Zhang, and Mengwei Xu.
\newblock Uishift: Enhancing vlm-based gui agents through self-supervised
  reinforcement learning.
\newblock \emph{arXiv preprint arXiv:2505.12493}, 2025.

\bibitem[Geirhos et~al.(2020)Geirhos, Jacobsen, Michaelis, Zemel, Brendel,
  Bethge, and Wichmann]{geirhos2020shortcut}
Robert Geirhos, J{\"o}rn-Henrik Jacobsen, Claudio Michaelis, Richard Zemel,
  Wieland Brendel, Matthias Bethge, and Felix~A Wichmann.
\newblock Shortcut learning in deep neural networks.
\newblock \emph{Nature Machine Intelligence}, 2\penalty0 (11):\penalty0
  665--673, 2020.

\bibitem[{Google}(2021)]{google2021android}
{Google}.
\newblock Android monkey.
\newblock \url{https://developer.android.com/studio/test/monkey}, 2021.
\newblock Accessed: 2026-01-03.

\bibitem[{Google}(2025)]{gemini3flash}
{Google}.
\newblock Gemini 3 flash, 2025.
\newblock URL \url{https://deepmind.google/models/gemini/flash/}.
\newblock Accessed: 2026-01-29.

\bibitem[Gou et~al.(2021)Gou, Yu, Maybank, and Tao]{gou2021knowledge}
Jianping Gou, Baosheng Yu, Stephen~J Maybank, and Dacheng Tao.
\newblock Knowledge distillation: A survey.
\newblock \emph{International journal of computer vision}, 129\penalty0
  (6):\penalty0 1789--1819, 2021.

\bibitem[Gu et~al.(2019)Gu, Sun, Ma, Cao, Xu, Yao, Zhang, Lu, and
  Su]{gu2019practical}
Tianxiao Gu, Chengnian Sun, Xiaoxing Ma, Chun Cao, Chang Xu, Yuan Yao, Qirun
  Zhang, Jian Lu, and Zhendong Su.
\newblock Practical {GUI} testing of {Android} applications via model
  abstraction and refinement.
\newblock In \emph{ICSE}, pages 269--280, 2019.

\bibitem[Ha and Schmidhuber(2018)]{ha2018world}
David Ha and J{\"u}rgen Schmidhuber.
\newblock World models.
\newblock \emph{arXiv preprint arXiv:1803.10122}, 2\penalty0 (3), 2018.

\bibitem[Hafner et~al.(2019)Hafner, Lillicrap, Ba, and
  Norouzi]{hafner2019dream}
Danijar Hafner, Timothy Lillicrap, Jimmy Ba, and Mohammad Norouzi.
\newblock Dream to control: Learning behaviors by latent imagination.
\newblock \emph{arXiv preprint arXiv:1912.01603}, 2019.

\bibitem[Hafner et~al.(2025)Hafner, Pasukonis, Ba, and
  Lillicrap]{hafner2025dreamerv3}
Danijar Hafner, Jurgis Pasukonis, Jimmy Ba, and Timothy Lillicrap.
\newblock Mastering diverse control tasks through world models.
\newblock \emph{Nature}, pages 1--7, 2025.

\bibitem[Hong et~al.(2024)Hong, Wang, Lv, Xu, Yu, Ji, Wang, Wang, Dong, Ding,
  et~al.]{hong2024cogagent}
Wenyi Hong, Weihan Wang, Qingsong Lv, Jiazheng Xu, Wenmeng Yu, Junhui Ji, Yan
  Wang, Zihan Wang, Yuxiao Dong, Ming Ding, et~al.
\newblock Cogagent: A visual language model for {GUI} agents.
\newblock In \emph{CVPR}, pages 14281--14290, 2024.

\bibitem[LeCun(2022)]{lecun2022path}
Yann LeCun.
\newblock A path towards autonomous machine intelligence version 0.9. 2,
  2022-06-27.
\newblock \emph{Open Review}, 62\penalty0 (1):\penalty0 1--62, 2022.

\bibitem[Li et~al.(2025)Li, Kallidromitis, Gokul, Kato, Kozuka, and
  Grover]{li2025mobileworldbench}
Shufan Li, Konstantinos Kallidromitis, Akash Gokul, Yusuke Kato, Kazuki Kozuka,
  and Aditya Grover.
\newblock Mobileworldbench: Towards semantic world modeling for mobile agents.
\newblock \emph{arXiv preprint arXiv:2512.14014}, 2025.

\bibitem[Li et~al.(2024)Li, Bishop, Li, Rawles, Campbell-Ajala, Tyamagundlu,
  and Riva]{li2024effects}
Wei Li, William~E Bishop, Alice Li, Christopher Rawles, Folawiyo
  Campbell-Ajala, Divya Tyamagundlu, and Oriana Riva.
\newblock On the effects of data scale on {UI} control agents.
\newblock \emph{NeurIPS}, 37:\penalty0 92130--92154, 2024.

\bibitem[Li et~al.(2020)Li, Li, He, Zheng, Li, and Guan]{li2020widget}
Yang Li, Gang Li, Luheng He, Jingjie Zheng, Hong Li, and Zhiwei Guan.
\newblock Widget captioning: Generating natural language description for mobile
  user interface elements.
\newblock \emph{arXiv preprint arXiv:2010.04295}, 2020.

\bibitem[Lin et~al.(2025{\natexlab{a}})Lin, Li, Gao, Yang, Wu, Bai, Lei, Wang,
  and Shou]{lin2025showui}
Kevin~Qinghong Lin, Linjie Li, Difei Gao, Zhengyuan Yang, Shiwei Wu, Zechen
  Bai, Stan~Weixian Lei, Lijuan Wang, and Mike~Zheng Shou.
\newblock Showui: One vision-language-action model for gui visual agent.
\newblock In \emph{Proceedings of the Computer Vision and Pattern Recognition
  Conference}, pages 19498--19508, 2025{\natexlab{a}}.

\bibitem[Lin et~al.(2025{\natexlab{b}})Lin, Liu, Lu, Yuan, Liu, Xu, Miao, Chao,
  and Li]{lin2025gui}
Musen Lin, Minghao Liu, Taoran Lu, Lichen Yuan, Yiwei Liu, Haonan Xu, Yu~Miao,
  Yuhao Chao, and Zhaojian Li.
\newblock Gui-rewalk: Massive data generation for gui agent via stochastic
  exploration and intent-aware reasoning.
\newblock \emph{arXiv preprint arXiv:2509.15738}, 2025{\natexlab{b}}.

\bibitem[Lu et~al.(2024)Lu, Shao, Liu, Meng, Li, Chen, Huang, Zhang, Qiao, and
  Luo]{lu2024gui}
Quanfeng Lu, Wenqi Shao, Zitao Liu, Fanqing Meng, Boxuan Li, Botong Chen,
  Siyuan Huang, Kaipeng Zhang, Yu~Qiao, and Ping Luo.
\newblock {GUI} odyssey: A comprehensive dataset for cross-app {GUI} navigation
  on mobile devices.
\newblock \emph{arXiv preprint arXiv:2406.08451}, 2024.

\bibitem[Luo et~al.(2025)Luo, Tang, Li, Papoudakis, Song, Gong, Hao, Wang, and
  Shao]{luo2025vimo}
Dezhao Luo, Bohan Tang, Kang Li, Georgios Papoudakis, Jifei Song, Shaogang
  Gong, Jianye Hao, Jun Wang, and Kun Shao.
\newblock Vimo: A generative visual gui world model for app agents.
\newblock \emph{arXiv preprint arXiv:2504.13936}, 2025.

\bibitem[Marion et~al.(2023)Marion, {\"U}st{\"u}n, Pozzobon, Wang, Fadaee, and
  Hooker]{marion2023less}
Max Marion, Ahmet {\"U}st{\"u}n, Luiza Pozzobon, Alex Wang, Marzieh Fadaee, and
  Sara Hooker.
\newblock When less is more: Investigating data pruning for pretraining llms at
  scale.
\newblock \emph{arXiv preprint arXiv:2309.04564}, 2023.

\bibitem[OpenAI(2025)]{openai2025gpt52systemcard}
OpenAI.
\newblock Update to gpt-5 system card: Gpt-5.2.
\newblock Technical report, OpenAI, 12 2025.
\newblock URL
  \url{https://cdn.openai.com/pdf/3a4153c8-c748-4b71-8e31-aecbde944f8d/oai_5_2_system-card.pdf}.
\newblock Accessed: 2026-01-29.

\bibitem[Pahuja et~al.(2025)Pahuja, Lu, Rosset, Gou, Mitra, Whitehead, Su, and
  Hassan]{pahuja2025explorer}
Vardaan Pahuja, Yadong Lu, Corby Rosset, Boyu Gou, Arindam Mitra, Spencer
  Whitehead, Yu~Su, and Ahmed Hassan.
\newblock Explorer: Scaling exploration-driven web trajectory synthesis for
  multimodal web agents.
\newblock In \emph{Findings of the Association for Computational Linguistics:
  ACL 2025}, pages 6300--6323, 2025.

\bibitem[Qin et~al.(2025)Qin, Ye, Fang, Wang, Liang, Tian, Zhang, Li, Li,
  Huang, et~al.]{qin2025ui}
Yujia Qin, Yining Ye, Junjie Fang, Haoming Wang, Shihao Liang, Shizuo Tian,
  Junda Zhang, Jiahao Li, Yunxin Li, Shijue Huang, et~al.
\newblock Ui-tars: Pioneering automated gui interaction with native agents.
\newblock \emph{arXiv preprint arXiv:2501.12326}, 2025.

\bibitem[Ramrakhya et~al.(2025)Ramrakhya, Szot, Attia, Yang, Nguyen, Mazoure,
  Gan, Agrawal, and Toshev]{ramrakhya2025scaling}
Ram Ramrakhya, Andrew Szot, Omar Attia, Yuhao Yang, Anh Nguyen, Bogdan Mazoure,
  Zhe Gan, Harsh Agrawal, and Alexander Toshev.
\newblock Scaling synthetic task generation for agents via exploration.
\newblock \emph{arXiv preprint arXiv:2509.25047}, 2025.

\bibitem[Ran et~al.(2023)Ran, Wang, Wang, and Xie]{ran2023badge}
Dezhi Ran, Hao Wang, Wenyu Wang, and Tao Xie.
\newblock Badge: prioritizing {UI} events with hierarchical multi-armed bandits
  for automated {UI} testing.
\newblock In \emph{ICSE}, pages 894--905, 2023.

\bibitem[Rawles et~al.(2023)Rawles, Li, Rodriguez, Riva, and
  Lillicrap]{rawles2023androidinthewild}
Christopher Rawles, Alice Li, Daniel Rodriguez, Oriana Riva, and Timothy
  Lillicrap.
\newblock Androidinthewild: A large-scale dataset for {Android} device control.
\newblock \emph{NeurIPS}, 36:\penalty0 59708--59728, 2023.

\bibitem[Shao et~al.(2024)Shao, Wang, Zhu, Xu, Song, Bi, Zhang, Zhang, Li, Wu,
  et~al.]{shao2024deepseekmath}
Zhihong Shao, Peiyi Wang, Qihao Zhu, Runxin Xu, Junxiao Song, Xiao Bi, Haowei
  Zhang, Mingchuan Zhang, YK~Li, Yang Wu, et~al.
\newblock Deepseekmath: Pushing the limits of mathematical reasoning in open
  language models.
\newblock \emph{arXiv preprint arXiv:2402.03300}, 2024.

\bibitem[Su et~al.(2017)Su, Meng, Chen, Wu, Yang, Yao, Pu, Liu, and
  Su]{su2017guided}
Ting Su, Guozhu Meng, Yuting Chen, Ke~Wu, Weiming Yang, Yao Yao, Geguang Pu,
  Yang Liu, and Zhendong Su.
\newblock Guided, stochastic model-based gui testing of android apps.
\newblock In \emph{Proceedings of the 2017 11th joint meeting on foundations of
  software engineering}, pages 245--256, 2017.

\bibitem[Sun et~al.(2025)Sun, Cheng, Ding, Jin, Wang, Xu, Wu, Jia, Chen, Liu,
  et~al.]{sun2025genesis}
Qiushi Sun, Kanzhi Cheng, Zichen Ding, Chuanyang Jin, Yian Wang, Fangzhi Xu,
  Zhenyu Wu, Chengyou Jia, Liheng Chen, Zhoumianze Liu, et~al.
\newblock Os-genesis: Automating gui agent trajectory construction via reverse
  task synthesis.
\newblock In \emph{Proceedings of the 63rd Annual Meeting of the Association
  for Computational Linguistics (Volume 1: Long Papers)}, pages 5555--5579,
  2025.

\bibitem[Tang et~al.(2025)Tang, Dong, Huang, Xiang, Ruan, Wang, Li, Xi, Cao,
  Pang, et~al.]{tang2025magicgui}
Liujian Tang, Shaokang Dong, Yijia Huang, Minqi Xiang, Hongtao Ruan, Bin Wang,
  Shuo Li, Zhiheng Xi, Zhihui Cao, Hailiang Pang, et~al.
\newblock Magicgui: A foundational mobile gui agent with scalable data pipeline
  and reinforcement fine-tuning.
\newblock \emph{arXiv preprint arXiv:2508.03700}, 2025.

\bibitem[{Tencent}(2024)]{WeChatMiniProgramDocs}
{Tencent}.
\newblock Weixin mini program platform capabilities, 2024.
\newblock URL
  \url{https://developers.weixin.qq.com/miniprogram/dev/platform-capabilities/miniapp/intro/}.
\newblock Accessed: 2026-01-29.

\bibitem[Torabi et~al.(2018)Torabi, Warnell, and Stone]{torabi2018behavioral}
Faraz Torabi, Garrett Warnell, and Peter Stone.
\newblock Behavioral cloning from observation.
\newblock \emph{arXiv preprint arXiv:1805.01954}, 2018.

\bibitem[Velasco and Roque(2025)]{velasco2025rethinking}
Dan~John Velasco and Matthew~Theodore Roque.
\newblock Rethinking the role of text complexity in language model pretraining.
\newblock In \emph{Proceedings of the First BabyLM Workshop}, pages 1--28,
  2025.

\bibitem[Wang et~al.(2025)Wang, Zou, Song, Feng, Fang, Lu, Liu, Luo, Liang,
  Huang, et~al.]{wang2025ui}
Haoming Wang, Haoyang Zou, Huatong Song, Jiazhan Feng, Junjie Fang, Junting Lu,
  Longxiang Liu, Qinyu Luo, Shihao Liang, Shijue Huang, et~al.
\newblock Ui-tars-2 technical report: Advancing gui agent with multi-turn
  reinforcement learning.
\newblock \emph{arXiv preprint arXiv:2509.02544}, 2025.

\bibitem[Wang* et~al.(2025)Wang*, Zhang*, Wang*, Gao*, Li*, Wang, Chen, Wan,
  Lu, Yang, Wang, Krishna, Wu, Fei-Fei, Choi, and Li]{wang2025vagen}
Kangrui Wang*, Pingyue Zhang*, Zihan Wang*, Yaning Gao*, Linjie Li*, Qineng
  Wang, Hanyang Chen, Chi Wan, Yiping Lu, Zhengyuan Yang, Lijuan Wang, Ranjay
  Krishna, Jiajun Wu, Li~Fei-Fei, Yejin Choi, and Manling Li.
\newblock Vagen:reinforcing world model reasoning for multi-turn vlm agents,
  2025.
\newblock URL \url{https://vagen-ai.github.io/}.

\bibitem[Wang et~al.(2024)Wang, Liu, Chen, Zhou, Gan, Zeng, Che, Yu, Hao, Shao,
  et~al.]{wang2024gui}
Shuai Wang, Weiwen Liu, Jingxuan Chen, Yuqi Zhou, Weinan Gan, Xingshan Zeng,
  Yuhan Che, Shuai Yu, Xinlong Hao, Kun Shao, et~al.
\newblock Gui agents with foundation models: A comprehensive survey.
\newblock \emph{arXiv preprint arXiv:2411.04890}, 2024.

\bibitem[Wei et~al.(2022)Wei, Wang, Schuurmans, Bosma, Xia, Chi, Le, Zhou,
  et~al.]{wei2022chain}
Jason Wei, Xuezhi Wang, Dale Schuurmans, Maarten Bosma, Fei Xia, Ed~Chi, Quoc~V
  Le, Denny Zhou, et~al.
\newblock Chain-of-thought prompting elicits reasoning in large language
  models.
\newblock \emph{Advances in neural information processing systems},
  35:\penalty0 24824--24837, 2022.

\bibitem[Wu et~al.(2024{\natexlab{a}})Wu, Wang, Ren, Cao, Li, Jiang, Ran, Hu,
  Yang, and Xie]{wu2024skill}
Mengzhou Wu, Hao Wang, Jun Ren, Yuan Cao, Yuetong Li, Alex Jiang, Dezhi Ran,
  Yitao Hu, Wei Yang, and Tao Xie.
\newblock Skill-adpative imitation learning for {UI} test reuse.
\newblock \emph{arXiv preprint arXiv:2409.13311}, 2024{\natexlab{a}}.

\bibitem[Wu et~al.(2024{\natexlab{b}})Wu, Wu, Xu, Wang, Sun, Jia, Cheng, Ding,
  Chen, Liang, et~al.]{wu2024atlas}
Zhiyong Wu, Zhenyu Wu, Fangzhi Xu, Yian Wang, Qiushi Sun, Chengyou Jia, Kanzhi
  Cheng, Zichen Ding, Liheng Chen, Paul~Pu Liang, et~al.
\newblock Os-atlas: A foundation action model for generalist gui agents.
\newblock \emph{arXiv preprint arXiv:2410.23218}, 2024{\natexlab{b}}.

\bibitem[Xu et~al.(2025{\natexlab{a}})Xu, Liu, Liu, Fu, Zhang, Jing, Zhang,
  Wang, Zhao, and Dong]{xu2025mobilerlonlineagenticreinforcement}
Yifan Xu, Xiao Liu, Xinghan Liu, Jiaqi Fu, Hanchen Zhang, Bohao Jing, Shudan
  Zhang, Yuting Wang, Wenyi Zhao, and Yuxiao Dong.
\newblock Mobilerl: Online agentic reinforcement learning for mobile gui
  agents, 2025{\natexlab{a}}.
\newblock URL \url{https://arxiv.org/abs/2509.18119}.

\bibitem[Xu et~al.(2025{\natexlab{b}})Xu, Wang, Wang, Lu, Xie, Saha, Sahoo, Yu,
  and Xiong]{xu2024aguvis}
Yiheng Xu, Zekun Wang, Junli Wang, Dunjie Lu, Tianbao Xie, Amrita Saha, Doyen
  Sahoo, Tao Yu, and Caiming Xiong.
\newblock Aguvis: Unified pure vision agents for autonomous {GUI} interaction.
\newblock In \emph{ICML}, 2025{\natexlab{b}}.

\bibitem[Ye et~al.(2025)Ye, Zhang, Xu, Liu, Wang, Zhu, Zheng, Gao, Cao, Lu,
  et~al.]{ye2025mobile}
Jiabo Ye, Xi~Zhang, Haiyang Xu, Haowei Liu, Junyang Wang, Zhaoqing Zhu, Ziwei
  Zheng, Feiyu Gao, Junjie Cao, Zhengxi Lu, et~al.
\newblock Mobile-agent-v3: Foundamental agents for gui automation.
\newblock \emph{arXiv preprint arXiv:2508.15144}, 2025.

\bibitem[Yue et~al.(2025)Yue, Chen, Lu, Zhao, Wang, Song, and
  Huang]{yue2025does}
Yang Yue, Zhiqi Chen, Rui Lu, Andrew Zhao, Zhaokai Wang, Shiji Song, and Gao
  Huang.
\newblock Does reinforcement learning really incentivize reasoning capacity in
  llms beyond the base model?
\newblock \emph{arXiv preprint arXiv:2504.13837}, 2025.

\bibitem[Zhang et~al.(2025{\natexlab{a}})Zhang, Neubig, and
  Yue]{zhang2025interplaypretrainingmidtrainingrl}
Charlie Zhang, Graham Neubig, and Xiang Yue.
\newblock On the interplay of pre-training, mid-training, and rl on reasoning
  language models, 2025{\natexlab{a}}.
\newblock URL \url{https://arxiv.org/abs/2512.07783}.

\bibitem[Zhang et~al.(2025{\natexlab{b}})Zhang, Zhang, Yang, Zhu, Zhao, Cao,
  Chen, and Yu]{zhang2025progrm}
Danyang Zhang, Situo Zhang, Ziyue Yang, Zichen Zhu, Zihan Zhao, Ruisheng Cao,
  Lu~Chen, and Kai Yu.
\newblock Progrm: Build better gui agents with progress rewards.
\newblock \emph{arXiv preprint arXiv:2505.18121}, 2025{\natexlab{b}}.

\bibitem[Zhang et~al.(2024)Zhang, Wu, Teng, Liao, Xu, Xiao, Wei, and
  Tang]{zhang2024android}
Jiwen Zhang, Jihao Wu, Yihua Teng, Minghui Liao, Nuo Xu, Xiao Xiao, Zhongyu
  Wei, and Duyu Tang.
\newblock Android in the zoo: Chain-of-action-thought for {GUI} agents.
\newblock \emph{arXiv preprint arXiv:2403.02713}, 2024.

\bibitem[Zhang et~al.(2025{\natexlab{c}})Zhang, Chen, Liu, Xue, Liao, Liu,
  Wang, Ning, Chen, Fu, et~al.]{zhang2025agent}
Kai Zhang, Xiangchao Chen, Bo~Liu, Tianci Xue, Zeyi Liao, Zhihan Liu, Xiyao
  Wang, Yuting Ning, Zhaorun Chen, Xiaohan Fu, et~al.
\newblock Agent learning via early experience.
\newblock \emph{arXiv preprint arXiv:2510.08558}, 2025{\natexlab{c}}.

\bibitem[Zhang et~al.(2025{\natexlab{d}})Zhang, Ni, Chen, Zhang, Rao, Peng, Lu,
  Hu, Guo, and Hu]{zhang2025bee}
Yi~Zhang, Bolin Ni, Xin-Sheng Chen, Heng-Rui Zhang, Yongming Rao, Houwen Peng,
  Qinglin Lu, Han Hu, Meng-Hao Guo, and Shi-Min Hu.
\newblock Bee: A high-quality corpus and full-stack suite to unlock advanced
  fully open mllms.
\newblock \emph{arXiv preprint arXiv:2510.13795}, 2025{\natexlab{d}}.

\end{thebibliography}

\newpage
\appendix
\newpage
\appendix
\onecolumn

\section{Extended Discussion on Related Work}
\label{app:related}

This appendix extends the discussion of background and related work in Section~\ref{sec:related}.

\subsection{Learning Paradigms for GUI Agents}
Recent advancements in generalist GUI agents have primarily focused on refining policy optimization paradigms.
While initial efforts relied on straightforward Behavioral Cloning (BC)~\cite{wu2024atlas, lin2025showui, lu2024gui, chen2024guicourse}, the field has rapidly shifted towards multi-step Reinforcement Learning (RL)~\cite{ye2025mobile, xu2025mobilerlonlineagenticreinforcement, wang2025ui} integrated with Chain-of-Thought (CoT)~\cite{wei2022chain} reasoning to handle non-stationary, long-horizon tasks.
However, recent studies on RL reveal a critical bottleneck: RL alone struggles to surpass the inherent capability ceiling of the base model~\cite{zhang2025interplaypretrainingmidtrainingrl, yue2025does}.
This underscores the necessity of elevating the foundation model's intrinsic capabilities via Continual Pre-Training (CPT) prior to policy alignment.
Existing CPT approaches for GUI agents, however, are largely confined to static UI understanding, such as element grounding or screen captioning~\cite{tang2025magicgui, wu2024atlas}.
While these methods improve perception, they fail to capture the causal mechanics of interaction.
Distinct from these static objectives, our work targets a foundational dynamic objective: the internalization of a ``World Model'' via CPT on atomic transitions.
By grounding the model in the forward dynamics of the interface (i.e., predicting consequences), we effectively elevate the capability ceiling of the base model, providing a scalable path to establishing this essential foundation for generalist GUI agents.

\subsection{Scalable Data Acquisition for GUI Agents}
A fundamental challenge to training generalist GUI agents is the data scalability bottleneck of Behavioral Cloning (BC).
Practically, collecting high-quality human demonstrations is prohibitively expensive.
To bypass these limitations, the field has pivoted toward synthetic data generation, primarily following two directions.
One line of work attempts to synthesize execution trajectories by deploying agents to interact with the environment under specific instructions~\cite{lin2025gui,sun2025genesis,ramrakhya2025scaling,pahuja2025explorer}.
However, due to the strict long-horizon dependencies of GUI tasks, generating valid data is notoriously difficult; a single intermediate error invalidates the entire trajectory, resulting in extremely low yield rates for high-quality samples.
Alternatively, other approaches focus on static tasks such as screen captioning or VQA by leveraging powerful foundation models to label GUI screenshots~\cite{qin2025ui,wu2024atlas, tang2025magicgui}.
Yet, this relies on Model-based Distillation, creating an inherent ``distillation ceiling'': the agent is bounded by the teacher model's hallucinations and lacks objective grounding in real execution.
In contrast, our \toolname{} framework shifts the source of supervision from fallible teacher models to intrinsic environmental dynamics.
By treating state transitions as objective verification, we overcome the fragility of long-horizon synthesis and the bias of distillation, ensuring that performance scales robustly with autonomous exploration.

\subsection{World Models and Dynamics in Digital Interfaces}
Conceptualizing the GUI as a dynamic system aligns with the broader literature on world models in model-based Reinforcement Learning and representation learning, where frameworks~\cite{hafner2025dreamerv3,assran2023self, lecun2022path} enable agents to plan by modeling future latent states.
In the specific domain of GUIs, prior works such as ViMo~\cite{luo2025vimo} and VAGEN~\cite{wang2025vagen} have explored this direction through explicit visual state prediction, while GUI-Shift~\cite{gao2025uishift} employs an inverse dynamics objective to infer actions from state changes.
Recent concurrent work~\cite{zhang2025agent} proposes utilizing implicit world modeling as an auxiliary task, yet remains constrained by local branching from expert demonstrations, lacking a scalable, task-agnostic Continual Pre-Training objective to master global environment dynamics.
We distinguish our work by leveraging autonomous exploration and intrinsic environmental feedback to internalize a GUI world model via Continual Pre-Training, establishing a robust dynamics foundation before agentic post-training.

\section{Prompt Templates}
\label{app:prompts}

We present the specific prompt templates utilized for the synthetic environmental dynamics Continual Pre-Training (CPT) of \toolname{} and the downstream GUI agent navigation tasks.

\subsection{Synthetic Environmental Dynamics Tasks}
\label{app:cpt_prompts}

The CPT tasks are designed to enable the model to master the underlying transition dynamics of the GUI environment. We adopt the following notations: $I_t$ denotes the screenshot at step $t$; $u_t$ denotes the natural language description of the action; $a_t$ denotes the atomic action (e.g., coordinates); and $\mathcal{D}_t$ denotes the textual description of the interface state.

\paragraph{Forward Dynamics ($I_t, u_t \to \mathcal{D}_{t+1}$ and $I_t, a_t \to \mathcal{D}_{t+1}$).}
This task requires the model to predict the future state description given the current image and an action.

\begin{tcolorbox}[title=Forward Dynamics Prompt Template]
\textbf{System:} You are a GUI operation expert. \\
\textbf{User:} $<$image$>$ Given the current image and the action performed on the image, predict the subsequent page in terms of its content and functionality. \\
\textbf{Input:} \{action\_summary\} or \{atomic\_action\}
\end{tcolorbox}

\paragraph{Inverse Dynamics ($I_t, I_{t+1} \to u_t / a_t$).}
This task infers the action performed between two states.

\begin{tcolorbox}[title=Inverse Dynamics Prompt Template]
\textbf{System:} You are a GUI operation expert. \\
\textbf{User:} $<$image$>$ $<$image$>$ Given the before and after images, predict the operation performed by the user. \\
\textbf{Constraint:} Describe the operation in natural language OR select from atomic actions (click $x$ $y$, input $x$ $y$ $t$, scroll $x$ $y$ $direction$).
\end{tcolorbox}

\paragraph{Inverse Dynamics with Goal ($I_t, \mathcal{D}_{t+1} \to u_t / a_t$).}
The agent infers the action required to reach a specific described state.

\begin{tcolorbox}[title=Inverse Dynamics (State Description) Prompt Template]
\textbf{System:} You are a GUI operation expert. \\
\textbf{User:} $<$image$>$ Given the current interface image, predict the operation performed by the user that would lead to the page described below. \\
\textbf{Target Description:} \{after\_screenshot\_summary\}
\end{tcolorbox}

\paragraph{Backward Dynamics ($u_t, I_{t+1} \to \mathcal{D}_t$).}
This task reconstructs the previous state description.

\begin{tcolorbox}[title=Backward Dynamics Prompt Template]
\textbf{System:} You are a GUI operation expert. \\
\textbf{User:} $<$image$>$ Given the resulting page image and the intermediate action description, predict the content and functionality of the previous page. \\
\textbf{Input:} \{action\_summary\}
\end{tcolorbox}

\subsection{GUI Navigation Task}
\label{app:nav_prompts}

For the downstream navigation task, we utilize a Chain-of-Thought (CoT) approach.

\begin{tcolorbox}[title=Navigation Prompt]
<image> <image> \\
You are an advanced GUI navigation assistant capable of perceiving and interacting with mobile interfaces. The provided image sequence represents the history of operations, with the last image showing the current UI state.

\vspace{0.5em}

\textbf{Current Instruction:} \{instruction\} \\
\textbf{User History:} \{history\}

\vspace{0.5em}

Based on the visual context and history, predict the next optimal operation. Your output must strictly adhere to the following XML-style format:

\texttt{<think>...</think><sub\_goal>...</sub\_goal><answer>...</answer>}

\vspace{0.5em}

\textbf{1. Thinking Process (\texttt{<think>})} \\
Conduct a step-by-step reasoning process: (1) Analyze the current UI layout and identify interactive elements; (2) Verify if previous actions were successful; (3) Reflect on any potential errors or deviations; (4) Formulate a high-level strategy to align with the user's instruction.

\vspace{0.5em}

\textbf{2. Sub-goal Planning (\texttt{<sub\_goal>})} \\
Translate your strategy into a concise, natural language description of the immediate next step (e.g., ``Click the search icon on the top right'', ``Scroll down to find the settings'').

\vspace{0.5em}

\textbf{3. Action (\texttt{<answer>})} \\
Map the sub-goal to a precise executable primitive from the following 5 categories:
\begin{enumerate}
    \setlength{\topsep}{0pt}
    \setlength{\partopsep}{0pt}
    \setlength{\itemsep}{0pt}
    \setlength{\parskip}{0pt}
    \setlength{\parsep}{0pt}
    
    \item \texttt{click x y} -- Tap at coordinates $(x, y)$.
    \item \texttt{input x y t} -- Focus at $(x, y)$ and type text $t$.
    \item \texttt{finish} -- Terminate the task successfully.
    \item \texttt{scroll x y dir} -- Scroll from $(x, y)$ in direction \texttt{dir}.
    \item \texttt{wait} -- Pause for system response.
\end{enumerate}
\end{tcolorbox}

\subsection{Generalization Evaluation Prompts}
\label{app:generalization_prompts}

To rigorously assess cross-domain transfer (Level 1) and compositional generalization (Level 2) as discussed in Section~\ref{subsec:generalization}, we employ specialized prompt templates that facilitate objective verification via a VLM judge.

\paragraph{Forward Dynamics Generalization (L1 \& L2).}
Unlike the standard CPT objective which generates a holistic description, the evaluation prompts require the model to explicitly predict 5 distinct UI elements.

\begin{tcolorbox}[title=Forward Dynamics L1 (Atomic) Evaluation Prompt]
\textbf{System:} You are a GUI operation expert. \\
\textbf{User:} $<$image$>$ Given the current image and an action description, predict the 5 key elements that will appear on the subsequent page.

\textbf{Action Description:} \{action\_description\}

Please output exactly five elements following the format below: \\
\texttt{<element\_1>Description of element 1, be concise</element\_1>} \\
\texttt{<element\_2>Description of element 2, be concise</element\_2>} \\
\texttt{<element\_3>Description of element 3, be concise</element\_3>} \\
\texttt{<element\_4>Description of element 4, be concise</element\_4>} \\
\texttt{<element\_5>Description of element 5, be concise</element\_5>}
\end{tcolorbox}

\begin{tcolorbox}[title=Forward Dynamics L2 (Compositional) Evaluation Prompt]
\textbf{System:} You are a GUI operation expert. \\
\textbf{User:} $<$image$>$ Given the current image and two consecutive action descriptions, predict the 5 key elements that will appear on the subsequent page after executing Action 1 on the current page, followed by Action 2 on the transitioned page.

\textbf{Action Description 1:} \{action\_description\_1\} \\
\textbf{Action Description 2:} \{action\_description\_2\}

Please output exactly five elements following the format below: \\
\texttt{<element\_1>Description of element 1, be concise</element\_1>} \\
\texttt{<element\_2>Description of element 2, be concise</element\_2>} \\
\texttt{<element\_3>Description of element 3, be concise</element\_3>} \\
\texttt{<element\_4>Description of element 4, be concise</element\_4>} \\
\texttt{<element\_5>Description of element 5, be concise</element\_5>}
\end{tcolorbox}

\paragraph{Inverse Dynamics Generalization (L2).}
For Level 2 Inverse Dynamics, the model must infer the initial action in a two-step chain, given only the start and end states (skipping the intermediate observation).

\begin{tcolorbox}[title=Inverse Dynamics L2 (Compositional) Evaluation Prompt]
\textbf{System:} You are a GUI operation expert. \\
\textbf{User:} $<$image$>$ $<$image$>$ Given the first and the third images, please describe in natural language the operation performed on the first image that causes it to generate an intermediate page, which is subsequently transformed into the third image via another operation?
\end{tcolorbox}
\subsection{VLM-based Verification Judges}
\label{app:vlm_judges}

To automate the evaluation of open-ended generation in our generalization experiments (Section~\ref{subsec:generalization}), we employ two distinct VLM judges. Both judges utilize a standardized structure to ensure rigorous and consistent verification.

\paragraph{Judge for Inverse Dynamics (Action Consistency).}
This judge evaluates whether the predicted action is visually and functionally equivalent to the ground truth (Binary Score: 0 or 1).

\begin{tcolorbox}[title=Inverse Dynamics Judge Prompt]
\textbf{System:} You are a professional UI automation judge and visual semantic evaluation expert.

\textbf{Task:} Determine whether the ``Model Predicted Action'' is visually and functionally consistent with the ``Ground Truth Action'' based on the provided UI screenshot.

\textbf{Input:}
\begin{itemize}
    \setlength{\itemsep}{0pt}
    \item \textbf{UI Screenshot:} $<$image$>$
    \item \textbf{Ground Truth Action:} \{ground\_truth\}
    \item \textbf{Model Predicted Action:} \{prediction\}
\end{itemize}

\textbf{Evaluation Criteria:}
\begin{itemize}
    \setlength{\itemsep}{0pt}
    \item \textbf{Visual Alignment:} The predicted action must target the same visual element (e.g., same icon/button) as the ground truth.
    \item \textbf{Action Compatibility:} The action type (e.g., Click, Scroll) must be consistent.
    \item \textbf{Flexibility:} Synonyms (e.g., ``Submit'' vs ``Confirm'') and unambiguous location-based descriptions are accepted.
\end{itemize}

\textbf{Output Format:} \\
\texttt{<reason>Brief analysis of visual alignment.</reason>} \\
\texttt{<score>1 (Pass) or 0 (Fail)</score>}
\end{tcolorbox}

\paragraph{Judge for Forward Dynamics (Element Prediction).}
This judge evaluates the accuracy of the predicted next-state elements by calculating the precision of the top-5 predictions (Discrete Score: 0, 0.2, ..., 1).

\begin{tcolorbox}[title=Forward Dynamics Judge Prompt]
\textbf{System:} You are a professional UI automation judge and visual semantic evaluation expert.

\textbf{Task:} Verify whether the UI elements listed in the ``Predicted Answer'' are actually present in the provided ``Reference Image'' (Ground Truth Next State).

\textbf{Input:}
\begin{itemize}
    \setlength{\itemsep}{0pt}
    \item \textbf{Reference Screenshot:} $<$image$>$
    \item \textbf{Predicted Answer:} \{predicted\_elements\}
\end{itemize}

\textbf{Evaluation Criteria:}
\begin{itemize}
    \setlength{\itemsep}{0pt}
    \item \textbf{Truncation Rule:} Evaluate \textbf{only the first 5 elements} in the predicted list. Ignore any subsequent elements.
    \item \textbf{Existence Check:} For each element, check if it is clearly visible in the Reference Screenshot.
    \item \textbf{Scoring Metric:} Calculate the score as $\frac{\text{Count of Existing Elements}}{5}$.
    \item \textbf{Score Set:} The final score must be exactly one of: \{0, 0.2, 0.4, 0.6, 0.8, 1\}.
\end{itemize}

\textbf{Output Format:} \\
\texttt{<reason>Briefly list which of the top-5 elements exist in the image.</reason>} \\
\texttt{<score>Score</score>}
\end{tcolorbox}

\section{Action Space Specification}
\label{app:action_space}

We define the action space $\mathcal{A}$ as a set of atomic operations supported by the agent. Each action is represented as a tuple $a_t = (\tau, \theta)$, where $\tau$ specifies the primitive type and $\theta$ contains the necessary parameters. 
The representation of spatial coordinates $(x, y)$ adapts to the underlying vision-language backbone:
\begin{itemize}
    \item \textbf{Qwen2.5-VL Series:} Utilizes \textbf{absolute pixel coordinates} $(x, y) \in [0, W] \times [0, H]$ based on the original screenshot resolution.
    \item \textbf{Qwen3-VL Series:} Utilizes \textbf{normalized coordinates} quantized to the integer range $[0, 1000]$.
\end{itemize}
The specific primitives and their parameter structures are summarized in Table~\ref{tab:action_space}.

\begin{table*}[h]
    \centering
    \caption{Definition of the Action Space.}
    \label{tab:action_space}
    \vspace{0.1in}
    
    \renewcommand{\arraystretch}{1.3}
    
    \begin{tabular}{l p{5.5cm} p{7cm}} 
    \toprule
    \textbf{Action Type ($\tau$)} & \textbf{Parameters ($\theta$)} & \textbf{Description} \\
    \midrule
    
    \texttt{CLICK} 
    & $(x, y)$ 
    \newline {\small Absolute pixels or normalized to $[0, 1000]$} 
    & Tap at the specified coordinates on the screen. \\
    \midrule
    
    \texttt{INPUT} 
    & $(x, y, \text{text})$ 
    & Focus on the target element at $(x, y)$ and type the provided text string. \\
    \midrule
    
    \texttt{SCROLL} 
    & $(x, y, \text{dir})$ \newline {\small $\text{dir} \in \{\text{UP, DOWN, LEFT, RIGHT}\}$} 
    & Perform a scroll gesture starting from $(x, y)$ towards the specified direction. \\
    \midrule
    
    \texttt{FINISH} 
    & $\emptyset$ 
    & Terminate the current episode immediately. \\
    \bottomrule
    \end{tabular}
\end{table*}

\begin{figure}[h]
    \centering
    \includegraphics[width=0.7\linewidth]{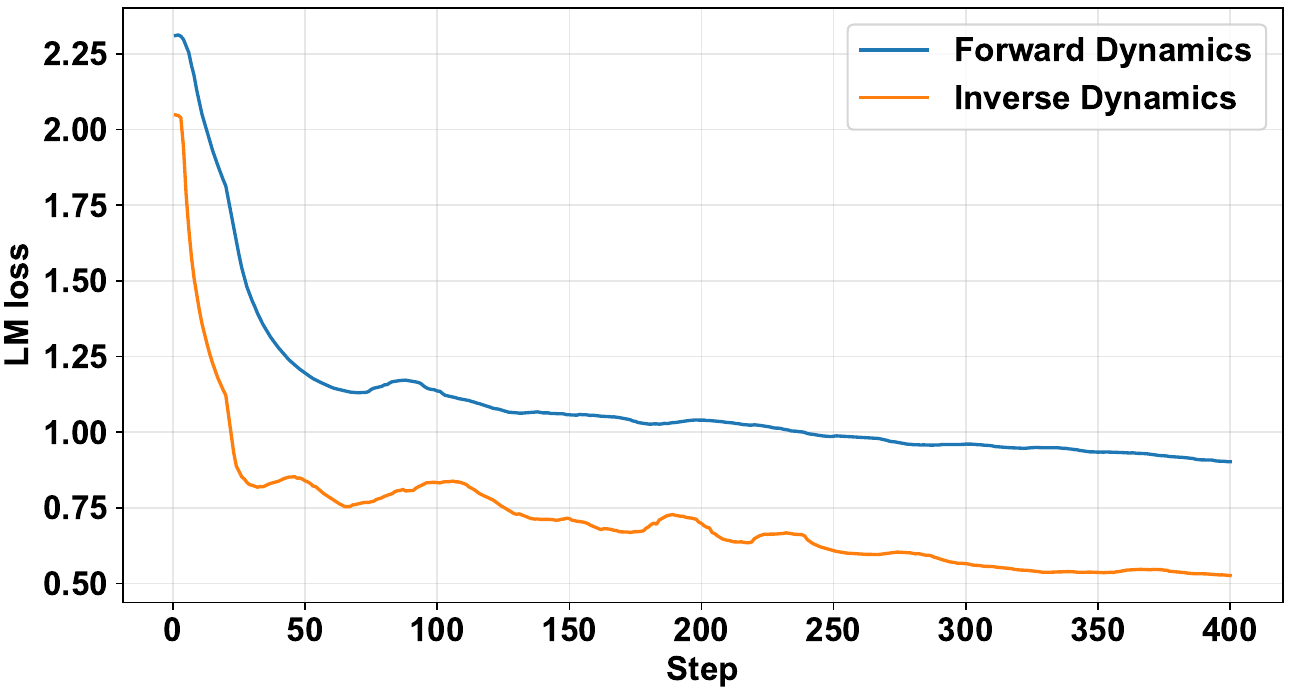}
    \caption{\textbf{Training Loss Comparison.} Inverse Dynamics (orange) exhibits rapid saturation, indicating insufficient task difficulty. In contrast, Forward Dynamics (blue) maintains a higher loss level, providing the sustained gradient signal necessary for effective representation learning.}
    \label{fig:loss_curve}
\end{figure}

\section{Details of Data Filtering Pipeline}
\label{app:data_filtering}

This section details the implementation of the deduplication stages using Locality-Sensitive Hashing (LSH) for both graph-structured accessibility trees and high-dimensional screenshots.

\subsection{Structural Transition Deduplication}
\label{app:structural_dedup}

To identify interface states with identical logic but varying content, we model transitions as sets of structural tokens.

\paragraph{Tokenization \& MinHash.}
We extract tokens from the accessibility tree nodes based on semantic attributes (e.g., $\texttt{tag}$, $\texttt{xpath}$, $\texttt{events}$). A transition is represented by the union of hashed tokens from both the pre- and post-action states. We generate MinHash signatures to efficiently estimate Jaccard similarity, reducing variable-length token sets to fixed-length vectors.

\paragraph{LSH Indexing.}
We employ a banding strategy to index these signatures. Transitions colliding in any band are retrieved as candidates. We then verify candidates against a Jaccard similarity threshold, retaining only unique representatives for the final dataset.

\subsection{Visual Transition Deduplication}
\label{app:visual_dedup}

For opaque components (e.g., WebViews), we utilize a composite visual fingerprinting strategy combining frequency and gradient domains.

\paragraph{Composite Hashing.}
For each transition, we concatenate the \textbf{Perceptual Hash (pHash)} and \textbf{Difference Hash (dHash)} of the start and end screenshots. This combination ensures robustness against rendering noise while remaining sensitive to significant layout changes.

\paragraph{Clustering Strategy.}
We employ bit-sampling LSH to project high-dimensional hashes into lower-dimensional buckets for indexing. Candidates are clustered using a Union-Find data structure based on Hamming distance thresholds. For each connected component, the transition with the highest data source priority is retained.

\subsection{Semantic Filtering}
We utilize Qwen3-VL-235B-Instruct as a semantic verifier. The model is prompted to output a binary validity score, retaining only transitions where the screen changes logically correspond to the annotated action.

\section{Training Dynamics Analysis}
\label{app:loss_analysis}

Figure~\ref{fig:loss_curve} illustrates the training dynamics discussed in Section~\ref{subsec:ablation}. 
The precipitous drop in Inverse Dynamics loss confirms that the task is computationally trivial, failing to provide meaningful gradient updates early in training. 
Conversely, Forward Dynamics imposes a sustained predictive challenge, driving continuous optimization of the world model's representations.

\section{Hyperparameter Settings}
\label{app:hyperparameters}

We detail the hyperparameter configurations for both the Continual Pre-Training (CPT) phase and the Agentic Post-Training phase.

\subsection{Continual Pre-Training (CPT)}
\label{app:cpt_hyperparams}

The hyperparameters for the CPT phase are summarized in Table~\ref{tab:cpt_hyperparams}. 
\begin{table}[h]
    \centering
    \caption{\textbf{Hyperparameters for Continual Pre-Training.}}
    \label{tab:cpt_hyperparams}
    \vspace{0.1in}
    \renewcommand{\arraystretch}{1.2}
    \setlength{\tabcolsep}{8pt}
    \begin{tabular}{l c c c}
    \toprule
    \textbf{Hyperparameter} & \textbf{32B Models} & \textbf{7B-8B Models} & \textbf{2B-4B Models} \\
    \midrule
    Optimizer & \multicolumn{3}{c}{AdamW} \\
    Global Batch Size & 2048 & 1024 & 1024 \\
    Peak Learning Rate & $1 \times 10^{-5}$ & $2 \times 10^{-5}$ & $3 \times 10^{-5}$ \\
    LR Scheduler & \multicolumn{3}{c}{Cosine} \\
    Warmup Ratio & \multicolumn{3}{c}{0.03} \\
    Max Sequence Length & \multicolumn{3}{c}{4096} \\
    Weight Decay & 0.1 & 0.01 & 0.01 \\
    Epochs & \multicolumn{3}{c}{1} \\
    \bottomrule
    \end{tabular}
\end{table}

\subsection{Agentic Post-Training: SFT}
\label{app:sft_hyperparams}

Table~\ref{tab:sft_hyperparams} lists the settings for the Supervised Fine-Tuning stage.

\begin{table}[h]
    \centering
    \caption{\textbf{Hyperparameters for SFT.}}
    \label{tab:sft_hyperparams}
    \vspace{0.1in}
    \renewcommand{\arraystretch}{1.2}
    \setlength{\tabcolsep}{8pt}
    \begin{tabular}{l c c c}
    \toprule
    \textbf{Hyperparameter} & \textbf{32B Models} & \textbf{7B-8B Models} & \textbf{2B-4B Models} \\
    \midrule
    Optimizer & \multicolumn{3}{c}{AdamW} \\
    Global Batch Size & \multicolumn{3}{c}{64} \\
    Peak Learning Rate & $3 \times 10^{-6}$ & $6 \times 10^{-6}$ & $1 \times 10^{-5}$ \\
    LR Scheduler & \multicolumn{3}{c}{Constant} \\
    Max Sequence Length & \multicolumn{3}{c}{4096} \\
    Weight Decay & \multicolumn{3}{c}{0} \\
    Epochs & \multicolumn{3}{c}{2} \\
    \bottomrule
    \end{tabular}
\end{table}

\section{SOTA Comparisons}
\label{app:sota_comparisons}

We evaluate the performance ceiling of our framework by comparing it against state-of-the-art proprietary models and large-scale open-weight baselines on both offline and online benchmarks.

\subsection{Offline Scalability Benchmark Comparison}
\label{app:offline_sota}

Table~\ref{tab:offline_sota} presents the performance comparison between our \toolname{} framework and state-of-the-art proprietary models (including GPT-5.2, Claude-4.5-Opus, and Gemini-3-Flash) as well as large-scale open-weight baselines on the offline scalability benchmark.

\begin{table}[h]
    \centering
    \caption{\textbf{Offline Benchmark Performance vs. Proprietary \& Large-Scale Models.} We report Exact Match (EM) and Type Match (TM) scores across different model families.}
    \label{tab:offline_sota}
    \vspace{0.1in}
    \renewcommand{\arraystretch}{1.2}
    \setlength{\tabcolsep}{10pt}
    \begin{tabular}{l c c}
    \toprule
    \textbf{Model} & \textbf{Exact Match (EM)} & \textbf{Type Match (TM)} \\
    \midrule
    \multicolumn{3}{l}{\textit{Proprietary Models}} \\
    GPT-5.2~\cite{openai2025gpt52systemcard} (w/ Grounding) & 61.1 & 76.9 \\
    Claude-4.5-Sonnet~\cite{anthropic2025claudesonnet45} (w/ Grounding) & 64.8 & 81.2 \\
    Claude-4.5-Opus~\cite{anthropic2025claudeopus45} (w/ Grounding) & \textbf{69.2} & 82.7 \\
    Gemini-3-Flash~\cite{gemini3flash} (w/ Grounding) & \textbf{69.2} & 82.0 \\
    \midrule
    \multicolumn{3}{l}{\textit{Open-Weight Baselines}} \\
    Qwen2.5-VL-32B & 33.9 & 56.6 \\
    Qwen2.5-VL-72B & 51.4 & 77.2 \\
    Qwen3-VL-32B & 47.1 & 79.2 \\
    Qwen3-VL-235B-A22B & 47.5 & 79.4 \\
    Qwen2.5-VL-32B (w/ Grounding) & 56.5 & 77.9 \\
    Qwen2.5-VL-72B (w/ Grounding) & 56.8 & 77.2 \\
    Qwen3-VL-32B (w/ Grounding) & 61.2 & 79.3 \\
    Qwen3-VL-235B-A22B (w/ Grounding) & 61.5 & 79.3 \\
    \midrule
    \multicolumn{3}{l}{\textit{Ours}} \\
    Qwen3-VL-32B + \toolname{} (CPT+SFT) & 65.0 & 81.1 \\
    Qwen2.5-VL-32B + \toolname{} (CPT+SFT) & 64.8 & 82.1 \\
    Qwen2.5-VL-32B + \toolname{} (CPT+SFT+GRPO) & 68.9 & \textbf{84.5} \\
    \bottomrule
    \end{tabular}
\end{table}

\subsection{Online Evaluation Benchmark Comparison}
\label{app:online_sota}

Complementing the offline metrics, Table~\ref{tab:online_sota} presents the Success Rate (SR) on the online benchmark. We list the performance of \toolname{} across three progressive post-training stages and compare them against representative external models, including state-of-the-art proprietary models.

\begin{table}[h]
    \centering
    \caption{\textbf{Online Success Rate vs. Proprietary Models.} Comparison of Success Rate (SR) on live mini-programs between our progressive pipeline and external baselines.}
    \label{tab:online_sota}
    \vspace{0.1in}
    \renewcommand{\arraystretch}{1.2}
    \setlength{\tabcolsep}{4pt} 
    \resizebox{\linewidth}{!}{%
        \begin{tabular}{l l c c}
        \toprule
        \textbf{Category} & \textbf{Model / Stage} & \textbf{SR (\%)} & \textbf{Gap w/ Best Ours} \\
        \midrule
        \multirow{3}{*}{\textit{Ours (32B)}} 
        & \toolname{} (SFT Cold Start) & 29.5 & \textcolor{red}{-14.8} \\
        & \hspace{1em} + Step-level GRPO & 38.9 & \textcolor{red}{-5.4} \\
        & \hspace{1em} \textbf{+ Multi-step GRPO} & \textbf{44.3} & -- \\
        \midrule
        \multirow{3}{*}{\shortstack[l]{\textit{Proprietary \&} \\ \textit{External}}} 
        & Seed-1.8~\cite{Seed1.8} (w/ Grounding) & 27.5 & \textcolor{red}{-16.8} \\
        & Claude-4.5-Opus~\cite{anthropic2025claudeopus45} (w/ Grounding) & 46.3 & \textcolor{blue}{+2.0} \\
        & Gemini-3-Flash~\cite{gemini3flash} (w/ Grounding) & 49.0 & \textcolor{blue}{+4.7} \\
        \bottomrule
        \end{tabular}%
    }
\end{table}

\end{document}